\documentclass{article}

\usepackage{microtype}
\usepackage{graphicx}
\usepackage{subfigure}
\usepackage{booktabs} % for professional tables
\usepackage{tikz}
\usepackage{amsfonts}
\usepackage{amsmath}
\usepackage{amssymb}
\usepackage{nicefrac}
\usepackage{siunitx}
\usepackage{textcomp}
\usepackage{url}
\usepackage{color}
\usepackage{xcolor,colortbl}
\usepackage{xspace}
\usepackage{times}
\usepackage{textcomp}
\usepackage{lipsum}
\newtheorem{theorem}{Theorem}
\usepackage[T1]{fontenc}
\usepackage{thmtools}
\usepackage{booktabs}
\usepackage{threeparttable}
\usepackage{lipsum}

\usepackage{hyperref}

\newcommand{\grayline}{\raisebox{2pt}{\tikz{\draw[-,black!40!gray,solid,line width = 0.9pt](0,0) -- (6mm,0);}}}

\newcommand{\blueline}{\raisebox{1pt}{\tikz{\draw[-,blue,dashed,line width = 1.5pt](0,0) -- (6mm,0);}}}
\newcommand{\greenline}{\raisebox{2pt}{\tikz{\draw[-,green,dashed,line width = 1.5pt](0,0) -- (6mm,0);}}}
\newcommand{\redline}{\raisebox{2pt}{\tikz{\draw[-,red,dashed,line width = 1.5pt](0,0) -- (6mm,0);}}}
\newcommand{\blackline}{\raisebox{2pt}{\tikz{\draw[-, black, dashed,line width = 1.5pt](0,0) -- (6mm,0);}}}
\newcommand{\purplerectangle}{\begin{tikzpicture}
    \filldraw[fill=blue] (0,0) -- (0.1,0.1732) -- (0.2,0) -- cycle;
    \end{tikzpicture}}
\newcommand{\mini}{\textit{mini}ImageNet}

\newcommand{\cifarfs}{\textsc{cifar-fs}}

\newcommand{\lstm}{\textsc{lstm}}
\def\abovestrut#1{\rule[0in]{0in}{#1}\ignorespaces}
\def\belowstrut#1{\rule[-#1]{0in}{#1}\ignorespaces}

\usepackage{amsmath,amsfonts,bm}

\def\eqref#1{equation~\ref{#1}}
\def\1{\bm{1}}

\DeclareMathAlphabet{\mathsfit}{\encodingdefault}{\sfdefault}{m}{sl}
\SetMathAlphabet{\mathsfit}{bold}{\encodingdefault}{\sfdefault}{bx}{n}

\newcommand{\softmax}{\mathrm{softmax}}

\newcommand{\KL}{D_{\mathrm{KL}}}

\DeclareMathOperator*{\argmin}{arg\,min}

\DeclareMathOperator{\Tr}{Tr}

\usepackage[accepted]{icml2020}

\icmltitlerunning{Learning to Learn Kernels with Variational Random Features}

\begin{document}

\twocolumn[
\icmltitle{Learning to Learn Kernels with Variational Random Features}




\icmlsetsymbol{equal}{*}

\begin{icmlauthorlist}
\icmlauthor{Xiantong Zhen}{equal,iiai,uva}
\icmlauthor{Haoliang Sun}{equal,sdu}
\icmlauthor{Yingjun Du}{equal,uva}
\icmlauthor{Jun Xu}{nku}
\icmlauthor{Yilong Yin}{sdu}
\icmlauthor{Ling Shao}{mbzuai,iiai}
\icmlauthor{Cees Snoek}{uva}
\end{icmlauthorlist}

\icmlaffiliation{iiai}{Inception Institute of Artificial Intelligence, UAE}
\icmlaffiliation{sdu}{School of Software, Shandong University, China}
\icmlaffiliation{uva}{Informatics Institute, University of Amsterdam, The Netherlands}
\icmlaffiliation{mbzuai}{Mohamed bin Zayed University of Artificial Intelligence, Abu Dhabi, UAE}
\icmlaffiliation{nku}{College of Computer Science, Nankai University, China}

% \icmlcorrespondingauthor{}
\icmlcorrespondingauthor{X. Zhen}{zhenxt@gmail.com}

\icmlkeywords{Machine Learning, ICML}

\vskip 0.3in
]



\printAffiliationsAndNotice{\icmlEqualContribution} % otherwise use the standard text.

\begin{abstract}
We introduce kernels with random Fourier features in the meta-learning framework for few-shot learning.
We propose meta variational random features (MetaVRF) to learn adaptive kernels for the base-learner, which is developed in a latent variable model by treating the random feature basis as the latent variable. We formulate the optimization of MetaVRF as a variational inference problem by deriving an evidence lower bound under the meta-learning framework. To incorporate shared knowledge from related tasks, we propose a context inference of the posterior, which is established by an LSTM architecture. The LSTM-based inference network effectively integrates the context information of previous tasks with task-specific information, generating informative and adaptive features. The learned MetaVRF is able to produce kernels of high representational power with a relatively low spectral sampling rate and also enables fast adaptation to new tasks. Experimental results on a variety of few-shot regression and classification tasks demonstrate that MetaVRF can deliver much better, or at least competitive, performance compared to existing meta-learning alternatives.
\end{abstract}

\section{Introduction}\label{sec:introduction}

Learning to learn, or \textit{meta-learning} \citep{Schmidhuber1992,thrun2012learning}, offers a promising tool for few-shot learning \citep{andrychowicz2016learning,ravi2017optimization,finn2017model} and has recently generated increasing popularity in machine learning. The crux of meta-learning for few-shot learning is to extract prior knowledge from related tasks to enable fast adaptation to a new task with a limited amount of data. Generally speaking, existing meta-learning algorithms~\citep{ravi2017optimization, bertinetto2018meta} design the meta-learner to extract meta-knowledge that improves the performance of the base-learner on individual tasks. Meta knowledge, like a good parameter initialization \citep{finn2017model}, or an efficient optimization update rule shared across tasks \citep{andrychowicz2016learning,ravi2017optimization} has been extensively explored in general learning framework, but how to define and use in few-shot learning remains an open question.

\begin{figure*}[t]
	\centering
	\includegraphics[width=.9\linewidth]{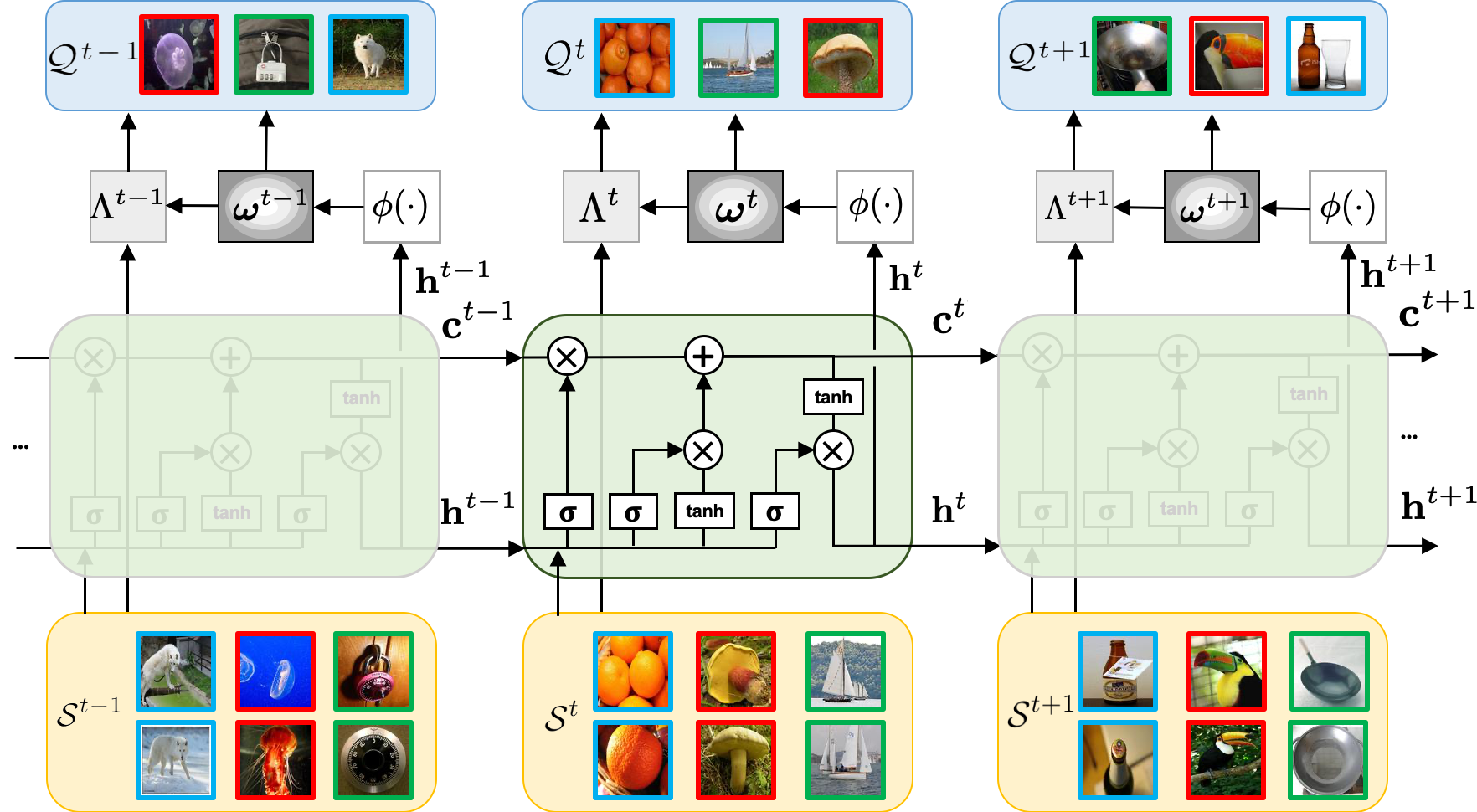}
\label{fig:frame}
	\vspace{-2mm}
	\caption{The learning framework of our meta variational random features (MetaVRF). The meta-learner employs an LSTM-based inference network $\phi(\cdot)$ to infer the spectral distribution over $\bm{\omega}^{t}$, the kernel from the support set $\mathcal{S}^t$ of the current task $t$, and the outputs $\mathbf{h}^{t-1}$ and $\mathbf{c}^{t-1}$ of the previous task. During the learning process, the cell state in the LSTM is deployed to accumulate the shared knowledge through experiencing a set of prior tasks. The \textit{remember} and \textit{forget} gates in the LSTM episodically refine the cell state by absorbing information from each experienced task. For each individual task, the task-specific information extracted from its support set is combined with distilled information from the previous tasks to infer the adaptive spectral distribution of the kernels.}
\end{figure*}

An effective base-learner should be powerful enough to solve individual tasks and able to absorb information provided by the meta-learner to improve its own performance. While potentially strong base-learners, kernels \citep{hofmann2008kernel} have not yet been studied in the meta-learning scenario for few-shot learning.  Learning adaptive kernels~\citep{bach2004multiple} in a data-driven way via random features~\citep{rahimi2008random} has demonstrated great success in regular learning tasks and remains of broad interest in machine learning~\citep{sinha2016learning,hensman2017variational,carratino2018learning,bullins2018not,li2019implicit}. However, due to the limited availability of data, it is challenging for few-shot learning to establish informative and discriminant kernels. We thus explore the relatedness among distinctive but relevant tasks to generate rich random features to build strong kernels for base-learners, while still maintaining their ability to adapt quickly to individual tasks.

In this paper, we make three important contributions. First, we propose meta variational random features (MetaVRF), integrating, for the first time, kernel learning with random features and variational inference into the meta-learning framework for few-shot learning. We develop MetaVRF in a latent variable model by treating the random Fourier basis of translation-invariant kernels as the latent variable. Second, we formulate the optimization of MetaVRF as a variational inference problem by deriving a new evidence lower bound (ELBO) in the meta-learning setting, where the posterior over the random feature basis corresponds to the spectral distribution associated with the kernel. This formulation under probabilistic modeling provides a principled way of learning data-driven kernels with random Fourier features and more importantly, fits well in the meta-learning framework for few-shot learning allowing us to flexibly customize the variational posterior to leverage the meta knowledge for inference.
As the third contribution, we propose a context inference which puts the inference of random feature bases of the current task into the context of all previous, related tasks. 
The context inference provides a generalized way to integrate context information of related tasks with task-specific information for the inference of random feature bases. To establish the context inference, we introduce a recurrent LSTM architecture~\citep{hochreiter1997long}, leveraging its innate capability of learning long-term dependencies, which can be adopted to explore shared meta-knowledge from a large set of previous tasks. The LSTM-based inference connects knowledge from previous tasks to the current task, gradually collecting and refreshing the knowledge across the course of learning. The learning process with an LSTM-based inference network is illustrated in Figure~\ref{fig:frame}. Once learning ceases, the ultimate LSTM state gains meta-knowledge from related experienced tasks, which enables fast adaptation to new tasks.

We demonstrate the effectiveness of the proposed MetaVRF by extensive experiments on a variety of few-shot regression and classification tasks. Results show that our MetaVRF achieves better, or at least competitive, performance compared to previous methods. Moreover, we conduct further analysis on MetaVRF to demonstrate its ability to be integrated with deeper architectures and its efficiency with relatively low sampling rates. We also apply MetaVRF to versatile and challenging settings with inconsistent training and test conditions, and it still delivers promising results, which further demonstrates its strong learning ability.

\section{Method}\label{sec:method}

We first describe the base-learner based on the kernel ridge regression in meta-learning for few-shot learning, and then introduce kernel learning with random features, based on which our meta variational random features are developed.

\subsection{Meta-Learning with Kernels}
We adopt the episodic training strategy commonly used for few-shot classification in meta-learning~\citep{ravi2017optimization}, which involves \textit{meta-training} and \textit{meta-testing} stages. In the \textit{meta-training} stage, a meta-learner is trained to enhance the performance of a base-learner on a \textit{meta-training} set with a batch of few-shot learning tasks, where a task is usually referred as an episode \citep{ravi2017optimization}. In the \textit{meta-test} stage, the base-learner is evaluated on a \textit{meta-testing} set with different classes of data samples from the \textit{meta-training} set.

For the few-shot classification problem, we sample $C$-way $k$-shot classification tasks from the \textit{meta-training} set, where $k$ is the number of labelled examples for each of the $C$ classes. Given the $t$-th task with a support set $\mathcal{S}^{t}=\{(\mathbf{x}_i, \mathbf{y}_i)\}_{i=1}^{C\mathord\times k}$ and query set $\mathcal{Q}^{t}=\{(\tilde{\mathbf{x}}_i, \tilde{\mathbf{y}}_i)\}_{i=1}^m$ ($\mathcal{S}^{t}, \mathcal{Q}^{t} \subseteq \mathcal{X}$), we learn the parameters $\alpha^{t}$ of the predictor $f_{\alpha^{t}}$ using a standard learning algorithm with kernel trick $\alpha^{t} = \Lambda(\Phi(X), Y)$, where $\mathcal{S}^{t} = \{X, Y\}$.\ Here, $\Lambda$ is the base-learner and $\Phi: \mathcal{X} \rightarrow \mathbb{R}^\mathcal{X}$ is a mapping function from $\mathcal{X}$ to a dot product space $\mathcal{H}$. The similarity measure $\mathtt{k}(\mathbf{x}, \mathbf{x}')=\langle\Phi(\mathbf{x}),\Phi(\mathbf{x}')\rangle$ is usually called a kernel~\citep{hofmann2008kernel}.

As in traditional supervised learning problems, the base-learner for the $t$-th single task can use a predefined kernel, e.g., radius base function, to map the input into a dot product space for efficient learning. Once the base-learner is obtained on the support set, its performance is evaluated on the query set by the following loss function:
\begin{equation}
\vspace{-3mm}
\sum_{(\tilde{\mathbf{x}}, \tilde{\mathbf{y}}) \in \mathcal{Q}^{t}} 
L \left(f_{\alpha^t} \big(\Phi(\tilde{\mathbf{x}} )\big), \tilde{\mathbf{y}}\right),
%%\vspace{-3mm}
\end{equation}
where $L(\cdot)$ can be any differentiable function, e.g., cross-entropy loss. In the meta-learning setting for few-shot learning, we usually consider a batch of tasks.\ Thus, the meta-learner is trained by optimizing the following objective function \textsl{w.r.t.} the empirical loss on $T$ tasks
\begin{equation}
 \begin{aligned}
\vspace{-3mm}
\sum^T_{t} \sum_{(\tilde{\mathbf{x}}, \tilde{\mathbf{y}} ) \in \mathcal{Q}^{t}} L\left(f_{\alpha^{t}}\big(\Phi^{t}(\tilde{\mathbf{x}})\big), \tilde{\mathbf{y}}\right), \text{s.t.} \,\ \alpha^{t} = \Lambda\left(\Phi^{t}(X), Y\right),
\label{obj}
\vspace{-2mm}
\end{aligned}   
\end{equation}
where $\Phi^t$ is the feature mapping function which can be obtained by learning a task-specific kernel $\mathtt{k}^t$ for each task $t$ with data-driven random Fourier features.

In this work, we employ kernel ridge regression, which has an efficient closed-form solution, as the base-learner $\Lambda$ for few-shot learning.\ The kernel value in the Gram matrix $K \in \mathbb{R}^{Ck\times Ck}$ is computed as $\mathtt{k}(\mathbf{x}, \mathbf{x}') = \Phi(\mathbf{x}) \Phi(\mathbf{x}')^{\top}$, where ``${\top}$'' is the transpose operation. The base-learner $\Lambda$ for a single task is obtained by solving the following objective \textsl{w.r.t.} the support set of this task,
\begin{equation}
\Lambda = \argmin_{\alpha} \Tr[(Y-\alpha K) (Y-\alpha K)^{\top}] + \lambda \Tr[\alpha K \alpha^{\top}],
\label{krg}
\end{equation}
which admits a closed-form solution 
\begin{equation}
\alpha = Y(\lambda \mathrm{I} + K)^{-1}
\label{closed}
\end{equation}
The learned predictor is then applied to samples in the query set $\tilde{X}$:
\begin{equation}
\hat{Y}=f_{\alpha}(\tilde{X})=\alpha \tilde{K}, 
\end{equation} 
Here, $\tilde{K} = \Phi(X)\Phi(\tilde{X})^\top\in \mathbb{R}^{Ck\times m}$, with each element as $\mathtt{k}(\mathbf{x}, \tilde{\mathbf{x}})$ between the samples from the support and query sets. Note that we also treat $\lambda$ in (\ref{krg}) as a trainable parameter by leveraging the meta-learning setting, and all these parameters are learned by the meta-learner.  

Rather than using pre-defined kernels, we consider learning adaptive kernels with random Fourier features in a data-driven way. Moreover, we leverage the shared knowledge by exploring dependencies among related tasks to learn rich features for building up informative kernels.

\subsection{Random Fourier Features}
Random Fourier features (RFFs) were proposed to construct approximate translation-invariant kernels using explicit feature maps~\citep{rahimi2008random}, based on Bochner's theorem~\citep{rudin1962fourier}.
\begin{theorem}[Bochner's theorem]~\citep{rudin1962fourier} A continuous, real valued, symmetric and shift-invariant function $\mathtt{k}(\mathbf{x},\mathbf{x}') = \mathtt{k}(\mathbf{x}-\mathbf{x}')$ on $\mathbb{R}^d$ is a positive definite kernel if and only if it is the Fourier transform $p(\bm{\omega})$ of a positive finite measure such that
\begin{align}
\mathtt{k}(\mathbf{x},\mathbf{x}') =& \int_{\mathbb{R}^d} e^{i\bm{\omega}^\top(\mathbf{x}-\mathbf{x}')}dp(\bm{\omega}) = \mathbb{E}_{\bm{\omega}}[\zeta_{\bm{\omega}}(\mathbf{x})\zeta_{\bm{\omega}}(\mathbf{x}')^*],  \nonumber \\   &\text{where} \,\,\,\, \zeta_{\bm{\omega}}(\mathbf{x}) = e^{i\bm{\omega}^\top \mathbf{x}}.
\end{align}
\end{theorem}
It is guaranteed that $\zeta_{\bm{\omega}}(\mathbf{x})\zeta_{\bm{\omega}}(\mathbf{x}’)^*$ is an unbiased estimation of $\mathtt{k}(\mathbf{x}, \mathbf{x}')$ with sufficient RFF bases $\{\bm{\omega}\}$ drawn from $p(\bm{\omega})$~\citep{rahimi2008random}.

For a predefined kernel, e.g., radius basis function (RBF), we use Monte Carlo sampling to draw bases from the spectral distribution, which gives rise to the explicit feature map:
\begin{equation}
\mathbf{z}(\mathbf{x}) = \frac{1}{\sqrt{D}} [\cos(\bm{\omega}_1^{\top} \mathbf{x} + b_1), \cdots, \cos(\bm{\omega}_D^{\top} \mathbf{x} + b_D)],
\label{rfs}
\end{equation}
where $\{\bm{\omega}_1, \cdots, \bm{\omega}_D\}$ are the random bases sampled from $p(\bm{\omega})$, and $[b_1, \cdots, b_D]$ are $D$ biases sampled from a uniform distribution with a range of $[0, 2\pi]$. 
Finally, the kernel values $\mathtt{k}(\mathbf{x}, \mathbf{x}')=\mathbf{z}(\mathbf{x})\mathbf{z}(\mathbf{x}')^{\top}$ in $K$ are computed as the dot product of their random feature maps with the same bases. 

\section{Meta Variational Random Features}
We introduce our MetaVRF using a latent variable model in which we treat random Fourier bases as latent variables inferred from data. 
Learning kernels with random Fourier features is tantamount to finding the posterior distribution over random bases in a data-driven way. It is naturally cast into a variational inference problem, where the optimization objective is derived from an evidence lower bound (ELBO) under the meta-learning framework.

\subsection{Evidence Lower Bound}
From a probabilistic perspective, under the meta-learning setting for few-shot learning, the random feature basis is obtained by maximizing the conditional predictive log-likelihood of samples from the query set $\mathcal{Q}$. 
\begin{align}
&\max_{p} \sum_{(\mathbf{x},\mathbf{y})\in \mathcal{Q}} \log  p(\mathbf{y} | \mathbf{x}, \mathcal{S}) \\ &= \max_{p} \sum_{(\mathbf{x},\mathbf{y})\in \mathcal{Q}} \log  \int p(\mathbf{y} |\mathbf{x},  \mathcal{S}, \bm{\omega})  p(\bm{\omega} | \mathbf{x},  \mathcal{S}) d\bm{\omega}.
\label{likeli}
\end{align}
We adopt a conditional prior distribution $p(\bm{\omega} | \mathbf{x},  \mathcal{S})$ over the base $\bm{\omega}$ as in the conditional variational auto-encoder (CVAE) \citep{sohn2015learning} rather than an uninformative prior \citep{kingma2013auto,rezende2014stochastic}. By depending on the input $\mathbf{x}$, we infer the bases that can specifically represent the data, while leveraging the context of the current task by conditioning on the support set $\mathcal{S}$.

In order to infer the posterior $p(\bm{\omega} | \mathbf{y},\mathbf{x}, \mathcal{S})$ over $\bm{\omega}$, which is generally intractable, we resort to using a variational distribution $q_{\phi}(\bm{\omega}| \mathcal{S})$ to approximate it, where the base is conditioned on the support set $\mathcal{S}$ by leveraging meta-learning. We obtain the variational distribution by minimizing the Kullback-Leibler (KL) divergence
\vspace{-2mm}
\begin{equation}
\KL[q_{\phi}(\bm{\omega}| \mathcal{S}) || p(\bm{\omega} | \mathbf{y}, \mathbf{x},  \mathcal{S})].
\label{kl}
\end{equation}
By applying the Bayes' rule to the posterior $p(\bm{\omega}|\mathbf{y},\mathbf{x},  \mathcal{S})$, we derive the ELBO as
\vspace{-2mm}
\begin{align}
\log  p(\mathbf{y} | \mathbf{x},  \mathcal{S}) \geq \,\,\,   &\mathbb{E}_{q_{\phi}(\bm{\omega}| \mathcal{S})} \log \, p(\mathbf{y} | \mathbf{x},  \mathcal{S}, \bm{\omega} ) \nonumber\\ &- \KL[q_{\phi}(\bm{\omega}|\mathcal{S}) || p(\bm{\omega} | \mathbf{x},  \mathcal{S})].
\vspace{-2mm}
\end{align}
The first term of the ELBO is the predictive log-likelihood conditioned on the observation $\mathbf{x}$, $ \mathcal{S}$ and the inferred RFF bases $\bm{\omega}$. Maximizing it enables us to make an accurate prediction for the query set by utilizing the inferred bases from the support set. The second term in the ELBO minimizes the discrepancy between the meta variational distribution $q_{\phi}(\bm{\omega}|\mathcal{S})$ and the meta prior $p(\bm{\omega} | \mathbf{x}, \mathcal{S})$, which encourages samples from the support and query sets to share the same random Fourier bases. The full derivation of the ELBO is provided in the supplementary material.

We now obtain the objective by maximizing the ELBO with respect to a batch of $T$ tasks:
\begin{align}
\vspace{-4mm}
\mathcal{L} = &\frac{1}{T} \sum_{t=1}^{T} \Big( \sum_{(\mathbf{x},\mathbf{y})\in \mathcal{Q}^{t}} \!\!\!\! \mathbb{E}_{q_{\phi}(\bm{\omega}^t| \mathcal{S}^t)} \log \, p(\mathbf{y} | \mathbf{x},\mathcal{S}^t, \bm{\omega}^t ) \nonumber\\ &- \KL[q_{\phi}(\bm{\omega}^t|\mathcal{S}^t) || p(\bm{\omega}^t | \mathbf{x}, \mathcal{S}^t)]  \Big),
\label{vi-obj-base} 
\vspace{-4mm}
\end{align}
where $\mathcal{S}^t$ is the support set of the $t$-th task associated with its specific bases $\{\bm{\omega}^t_{d}\}_{d=1}^{D}$ and $(\mathbf{x}, \mathbf{y}) \in \mathcal{Q}^t$ is the sample from the query set of the $t$-th task. Directly optimizing the above objective does not take into account the task dependency. Thus, we introduce context inference by conditioning the posterior on both the support set of the current task and the shared knowledge extracted from previous tasks.

\begin{figure}[t]
	\centering
	\includegraphics[width=0.9\linewidth]{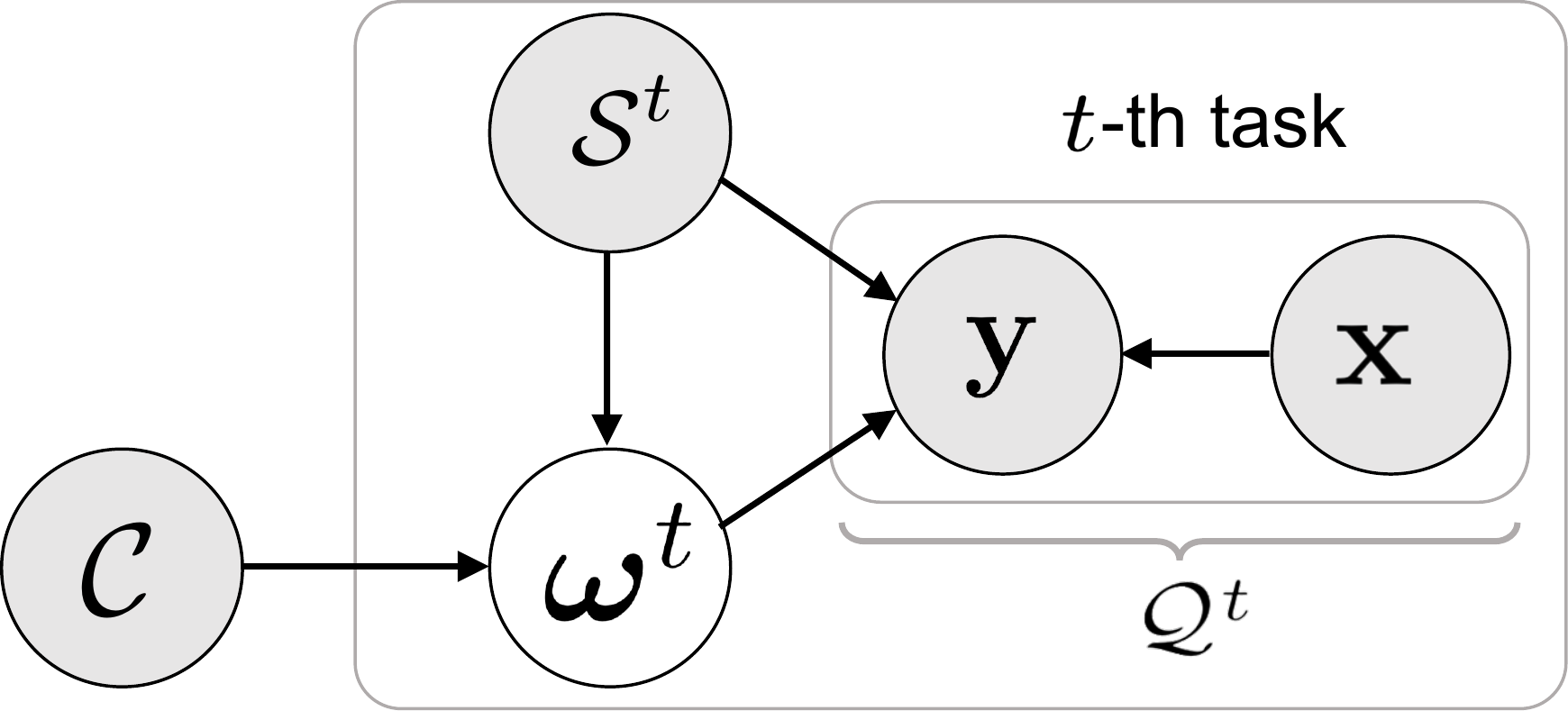}
	\caption{Illustration of MetaVRF in a directed graphical model, where $(\mathbf{x}, \mathbf{y})$ is a sample in the query set $\mathcal{Q}^t$. The base $\bm{\omega}^t$ of the $t$-th task is dependent on the support set $\mathcal{S}^t$ of the current task and the context $\mathcal{C}$ of related tasks. \label{graph}}
\end{figure}

\subsection{Context Inference}
We propose a context inference which puts the inference of random feature bases for the current task in the context of related tasks. We replace the variational distribution in (\ref{kl}) with a conditional distribution $q_{\phi}(\bm{\omega}^t| \mathcal{S}^t,\mathcal{C})$ that makes the bases $\{\bm{\omega}^t_{d}\}_{d=1}^{D}$ of the current $t$-th task conditioned also on the context $\mathcal{C}$ of related tasks. 

The context inference gives rise to a new ELBO, as follows:
\begin{equation}
\begin{aligned}
\log  p(\mathbf{y} | \mathbf{x},  \mathcal{S}^t) \geq \,\,\,   &\mathbb{E}_{q_{\phi}(\bm{\omega}| \mathcal{S}^t,\mathcal{C})} \log \, p(\mathbf{y} | \mathbf{x},  \mathcal{S}^t, \bm{\omega} ) \\ &- \KL[q_{\phi}(\bm{\omega}|\mathcal{S}^t,\mathcal{C}) || p(\bm{\omega} | \mathbf{x},  \mathcal{S}^t)],
\label{metaelbo} 
\end{aligned}
\end{equation}
which can be represented in a directed graphical model as shown in Figure~\ref{graph}. In a practical sense, the KL term in (\ref{metaelbo}) encourages the model to extract useful information from previous tasks for inferring the spectral distribution associated with each individual sample $\mathbf{x}$ of the query set in the current task. 

The context inference integrates the knowledge shared across tasks with the task-specific knowledge to build up adaptive kernels for individual tasks. The inferred random features are highly informative due to the absorbed information from prior knowledge of experienced tasks. The base-learner built on the inferred kernel with the informative random features can effectively solve the current task. %The variational posterior in (\ref{metaelbo}) offers a general formula to simultaneously incorporate shared and task-specific knowledge into the inference of kernels for each of individual tasks.

However, since there is usually a huge number of related tasks, it is non-trivial to model all these tasks simultaneously. We consider using recurrent neural networks to gradually accumulate information episodically along with the learning process by organizing tasks in a sequence. We propose an LSTM-based inference network by leveraging its innate capability of remembering long-term information~\citep{gers2000recurrent}. The LSTM offers a well-suited structure to implement the context inference. The cell state $\mathbf{c}$ stores and accrues the meta knowledge shared among related tasks, which can also be updated when experiencing a new task in each episode during the course of learning; the output $\mathbf{h}$ is used to adapt to each specific task. 

To be more specific, we model the variational posterior $q_{\phi}(\bm{\omega}^t| \mathcal{S}^t,\mathcal{C})$ through $q_{\phi}(\bm{\omega}|\mathbf{h}^t)$ which is parameterized as a multi-layer perceptron (MLP) $\phi(\mathbf{h}^t)$. Note that $\mathbf{h}^t$ is the output from an LSTM that takes $\mathcal{S}^t$ and $\mathcal{C}$ as inputs. We implement the inference network with both vanilla and bidirectional LSTMs \citep{schuster1997bidirectional,graves2005framewise}. For a vanilla LSTM, we have
\begin{equation}
[\mathbf{h}^t, \mathbf{c}^t] = g_{\mathrm{LSTM}}(\mathcal{\bar{S}}^t,\mathbf{h}^{t-1},\mathbf{c}^{t-1}),
\label{vlstm}
\end{equation}
where $g_{\mathrm{LSTM}}(\cdot)$ is a vanilla LSTM network that takes the current support set, the output $\mathbf{h}^{t-1}$ and the cell state $\mathbf{c}^{t-1}$ as the input. $\mathcal{\bar{S}}^t$ is the average over the feature representation vectors of samples in the support set~\citep{zaheer2017deep}. The feature representation is obtained by a shared convolutional network $\psi(\cdot)$. To incorporate more context information, we also implement the inference with a bidirectional LSTM, and we have $\mathbf{h}^t = [\stackrel{\rightarrow}{\mathbf{h}^t}, \stackrel{\leftarrow}{\mathbf{h}^t}]$, 
where $\stackrel{\rightarrow}{\mathbf{h}^t}$ and $\stackrel{\leftarrow}{\mathbf{h}^t}$ are the outputs from forward and backward LSTMs, respectively, and $[\cdot,\cdot]$ indicates a concatenation operation.

Therefore, the optimization objective with the context inference is:
\begin{equation}
\begin{aligned}
\mathcal{L} = &\frac{1}{T} \sum_{t=1}^{T} \Big( \sum_{(\mathbf{x},\mathbf{y})\in \mathcal{Q}^{t}} \!\!\!\! \mathbb{E}_{q_{\phi}(\bm{\omega}^t| \mathbf{h}^t)} \log \, p(\mathbf{y} | \mathbf{x},\mathcal{S}^t, \bm{\omega}^t) \\ &- \KL[q_{\phi}(\bm{\omega}^t|\mathbf{h}^t) || p(\bm{\omega}^t | \mathbf{x},\mathcal{S}^t)]  \Big).
\label{vi-obj}%\mathop{
\end{aligned}
\vspace{-2mm}
\end{equation}
where the variational approximate posterior $q_{\phi}(\bm{\omega}^t| \mathbf{h}^t)$ is taken as a multivariate Gaussian with a diagonal co-variance. Given the support set as input, the mean $\bm{\omega}_{\mu}$ and standard deviation $\bm{\omega}_{\sigma}$ are output from the inference network $\phi(\cdot)$. The conditional prior $p(\bm{\omega}^t | \mathbf{x},\mathcal{S}^t)$ is implemented with a prior network which takes an aggregated representation by using the cross attention \citep{kim2019attentive} between $\mathbf{x}$ and $\mathcal{S}^t$. The details of the prior network are provided in the supplementary material. To enable back-propagation with the sampling operation during training, we adopt the reparametrization trick \citep{rezende2014stochastic,kingma2013auto} as
$\bm{\omega}= \bm{\omega}_{\mu} + \bm{\omega}_{\sigma} \odot \boldsymbol\epsilon$, where $\bm\epsilon \sim \mathcal{N}(0, \mathrm{I} ).$

During the course of learning, the LSTMs accumulate knowledge in the cell state by updating their cells using information extracted from each task. For the current task $t$, the knowledge stored in the cell is combined with the task-specific information from the support set to infer the spectral distribution for this task. To accrue the information across all the tasks in the meta-training set, the output and the cell state of LSTMs are passed down across batches. As a result, the finial cell state contains the distilled prior knowledge from all those experienced tasks in the meta-training set. 

\textbf{Fast Adaptation.} Once meta-training ceases, the output and the cell state are directly used for a new incoming task in the meta-test set to achieve fast adaptation with a simple feed-forward computation operation. To be more specific, for a task with the support set $\mathcal{S}^*$ in the meta-test set, we draw $D$ samples $\{\bm{\omega}^{(l)}\}_{l=1}^D$ as the bases: $\bm{\omega}^{(l)}\sim q(\bm{\omega}|\mathbf{h}^*)$, where $\mathbf{h}^*$ is output from either a vanilla LSTM or a bidirectional LSTM, depending on which is used during the meta-training stage. The bases are adopted to compute the kernels on the support set and construct the classifier of the base-learner for the task, using (\ref{closed}). The classifier is then used to make predictions of samples in the query set for performance evaluation.

\section{Related Work}\label{sec:related}

\textbf{Meta-learning}, or learning to learn, endues machine learning models the ability to improve their performance by leveraging knowledge extracted from a number of prior tasks. It has received increasing research interest with breakthroughs in many directions~\citep{finn2017model,rusu2018meta,gordon2018meta,rajeswaran2019meta}. Gradient-based methods (e.g., MAML~\citep{finn2017model}) learn an appropriate initialization of model parameters and adapt it to new tasks with only a few gradient steps~\citep{finn2018, zintgraf2019fast, rusu2018meta}. Learning a shared optimization algorithm has also been explored in order to quickly learn of new tasks~\citep{ravi2017optimization,andrychowicz2016learning,chen2017learning}. 

%Memory-based methods learn to leverage an external memory module to store and leverage key knowledge for quick adaptation~\citep{santoro2016meta,ramalho2019adaptive}. 
%Bayesian meta-learning methods~\citep{edwards2016towards,finn2018probabilistic, gordon2018meta,saemundsson2018meta} usually rely on hierarchical Bayesian models to learn the shared statistical information among different tasks and reason about the uncertainty over models. 
%Differentiable solution methods~\citep{liu2018learning, bertinetto2018meta} learn a universal feature embedding and obtain a task-specific learner with a closed-form solution.

\textbf{Metric learning} has been widely studied with great success for few-shot learning~\citep{vinyals2016matching,snell2017prototypical,satorras2018few,oreshkin2018tadam,allen2019infinite}. The basic assumption is that a common metric space is shared across related tasks. By extending the matching network~\cite{vinyals2016matching} to few-shot scenarios, Snell et al.~\cite{snell2017prototypical} constructed a prototype for each class by averaging feature representations of samples from the class in the metric space. The classification is conducted via matching the query samples to prototypes by computing their distances. To enhance the prototype representation, Allen et al.~\cite{allen2019infinite} proposed an infinite mixture of prototypes (IMP) to adaptively represent data distribution of each class by multiple clusters instead of using a single vector. In addition, Oreshkin et al.~\cite{oreshkin2018tadam} proposed task dependent adaptive metric for improved few-shot learning and established prototypes of classes conditioning on a task representation encoded by a task embedding network. Their results indicate the benefit of learning task-specific metric in few-shot learning. 

%Graphical neural network (GNN) based models generalize the matching methods by learning the message propagation from the support set and transferring it to the query set~\citep{garcia2018few}.

While these meta-learning algorithms have made great progress in few-shot learning tasks, exploring prior knowledge from previous tasks remains an open challenge~\citep{titsias2019functional}. In this work, we introduce kernels based on random features as the base-learners, which enables us to acquire shared knowledge across tasks by modeling their dependency via the random feature basis of kernels.

\textbf{Kernel learning with random Fourier features} is a versatile and powerful tool in machine learning~\citep{bishop2006pattern, hofmann2008kernel, shervashidze2011weisfeiler}. Pioneering works~\citep{bach2004multiple,gonen2011multiple, duvenaud2013structure} learn to combine predefined kernels in a multi-kernel learning manner. Kernel approximation by random Fourier features (RFFs)~\citep{rahimi2008random} is an effective technique for efficient kernel learning~\citep{gartner2002multi}, which has recently become increasingly popular~\citep{sinha2016learning,carratino2018learning}. Recent works~\citep{wilson2013gaussian} learn kernels in the frequency domain by modeling the spectral distribution as a mixture of Gaussians and computing its optimal linear combination. Instead of modeling the spectral distribution with explicit density functions, other works focus on optimizing the random base sampling strategy~\citep{yang2015carte, sinha2016learning}. Nonetheless, it has been shown that accurate approximation of kernels does not necessarily result in high classification performance \citep{avron2016quasi,chang2017data}. This suggests that learning adaptive kernels with random features by data-driven sampling strategies \citep{sinha2016learning} can improve the performance, even with a low sampling rate compared to using universal random features \citep{avron2016quasi,chang2017data}.

Our MetaVRF is the first work to introduce kernel learning with random features to the meta-learning framework for few-shot learning. The optimization of MetaVRF is naturally cast as a variational inference and the context inference offers a principled way to incorporate prior knowledge and achieve informative and adaptive kernels.

\section{Experiments}\label{sec:experiments}

We evaluate our MetaVRF on several few-shot learning problems for both regression and classification. 
We demonstrate the benefit of exploring task dependency by implementing a baseline MetaVRF (\ref{vi-obj-base}) without using the LSTM, which infers the random base solely from the support set. We also conduct further analysis to validate the effectiveness of our MetaVRF by showing its performance with deep embedding architectures, different numbers of bases, and under versatile and challenging settings with inconsistent training and test conditions.

\begin{figure}[]
	\centerline{\includegraphics[width=1\linewidth]{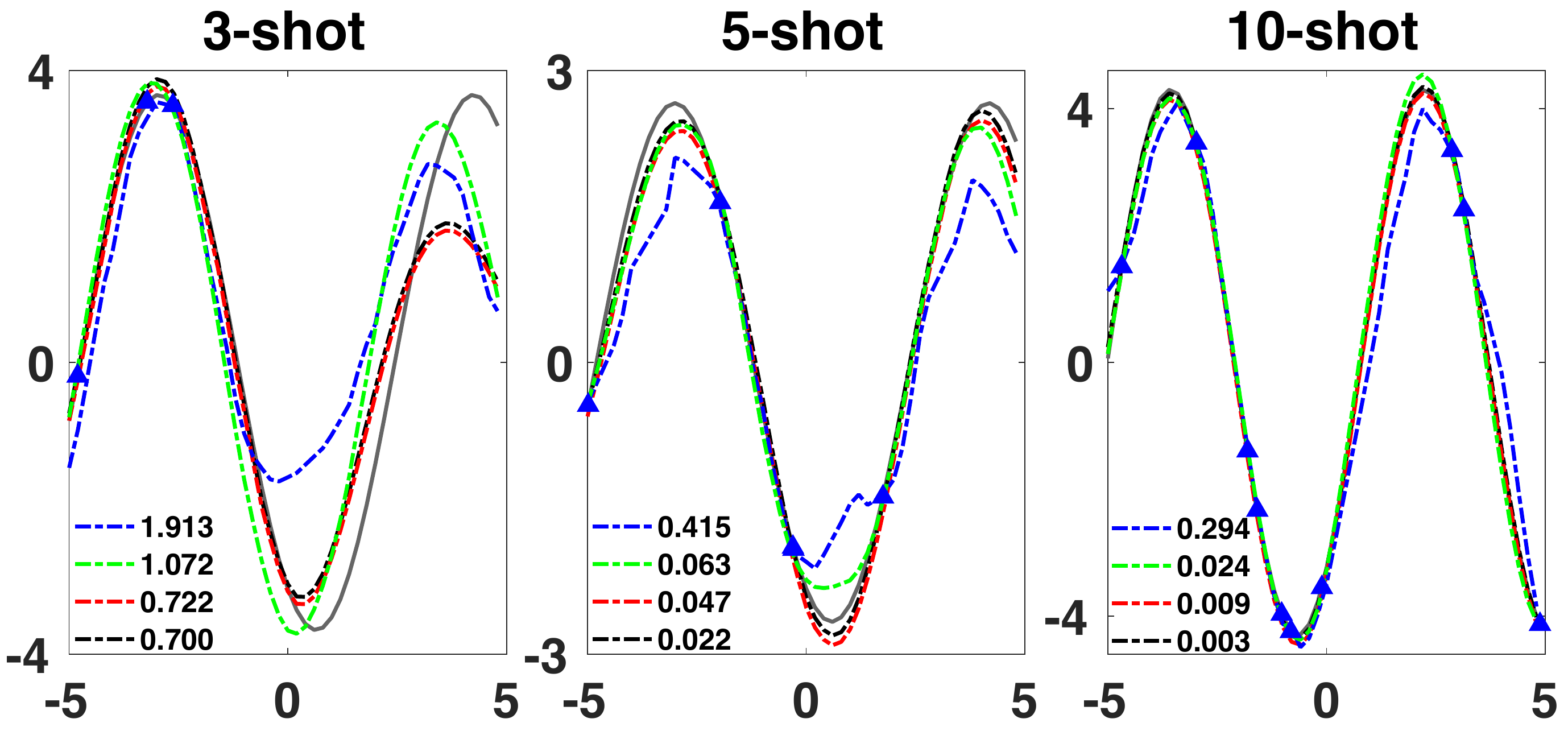}}
\vspace{-3mm}
	\caption{Performance (MSE) comparison for few-shot regression. Our MetaVRF fits the target function well, even with only three shots, and consistently outperforms regular RFFs and the counterpart MAML. 
	(\blackline~MetaVRF~with~bi-\lstm; \redline~MetaVRF~with~\lstm; \greenline~MetaVRF~w/o~\lstm; \blueline~MAML;~\grayline~Ground Truth; \purplerectangle~Support Samples.)}
\label{fig:reg} 
\end{figure}

\begin{table*}[t]
\small
\caption{Performance ($\%$) on \mini{} and \cifarfs{}.}\label{tab:miniandcifar}
\centering
%\begin{center}
%\begin{footnotesize}
%\begin{scriptsize}
\begin{tabular}{lcccc}
%\vspace{-0.05in}
\toprule
% \abovespace
& \multicolumn{2}{c}{\textbf{\mini{}, 5-way}} & \multicolumn{2}{c}{\textbf{\cifarfs{}, 5-way}} \\
% \belowspace
\textbf{Method} & 1-shot & 5-shot & 1-shot & 5-shot \\
\midrule
% \abovespace
%\textbf{\textsc{Matching net}}~\citep{vinyals2016matching} & 41.2  & 56.2  & --- & --- \\
\textbf{\textsc{Matching net}}~\citep{vinyals2016matching} & 44.2  & 57  & --- & --- \\
\textbf{\textsc{MAML}}~\citep{finn2017model} & 48.7$\pm$1.8  & 63.1$\pm$0.9  & 58.9$\pm$1.9  & 71.5$\pm$1.0  \\
\textbf{\textsc{MAML}} ($64$C) & 46.7$\pm$1.7  & 61.1$\pm$0.1  & 58.9$\pm$1.8  & 71.5$\pm$1.1  \\
\textbf{\textsc{Meta-LSTM}}~\citep{ravi2017optimization} & 43.4$\pm$0.8  & 60.6$\pm$0.7  & --- & --- \\
\textbf{\textsc{Proto net}}~\citep{snell2017prototypical} & 47.4$\pm$0.6  & 65.4$\pm$0.5  & 55.5$\pm$0.7  & 72.0$\pm$0.6  \\
%\textbf{\textsc{Proto net} $*$} & 42.9$\pm$0.6  & 65.9$\pm$0.6  & 57.9$\pm$0.8  & 76.7$\pm$0.6  \\
\textbf{\textsc{Relation net}}~\citep{sung2018learning} & 50.4$\pm$0.8  & 65.3$\pm$0.7  & 55.0$\pm$1.0  & 69.3$\pm$0.8  \\
%\textbf{\textsc{SNAIL}} (with \emph{ResNet})~\citep{mishra2018simple} & \emph{55.7$\pm$1.0 } & \emph{68.9$\pm$0.9 } & --- & --- \\
%\textbf{\textsc{SNAIL}} (64C)~\citep{mishra2018simple} & 48.7  & 60.4  & 56.3 & 70.5 \\
\textbf{\textsc{SNAIL}} (32C)~by \citep{bertinetto2018meta} & 45.1  & 55.2  & --- & --- \\
\textbf{\textsc{GNN}}~\citep{garcia2018few} & 50.3  & 66.4  & 61.9  & 75.3  \\
\textbf{\textsc{PLATIPUS}}~\citep{finn2018probabilistic} & 50.1$\pm$1.9  & --- & --- & --- \\
\textbf{\textsc{VERSA}}~\citep{gordon2018meta} & 53.3$\pm$1.8  & 67.3$\pm$0.9  & 62.5$\pm$1.7  & 75.1$\pm$0.9  \\
\textbf{R2-D2} ($64$C)~\citep{bertinetto2018meta} & 49.5$\pm$0.2  & 65.4$\pm$0.2  & 62.3$\pm$0.2  & \textbf{77.4}$\pm$0.2  \\
\textbf{\textsc{R2-D2}}~\citep{devosreproducing} & 51.7$\pm$1.8  & 63.3$\pm$0.9  & 60.2$\pm$1.8  & 70.9$\pm$0.9  \\
\textbf{\textsc{CAVIA}}~\citep{zintgraf2019fast} & 51.8$\pm$0.7  & 65.6$\pm$0.6  & --- & --- \\
\textbf{\textsc{iMAML}}~\citep{rajeswaran2019meta} & 49.3$\pm$1.9  &--- & --- & --- \\
%\textbf{\textsc{GNN}$*$} & 50.3  & \textbf{68.2}  & 56.0  & 72.5  \\
\midrule
%\textbf{\textsc{Ours/\rr{}}} (with 64C) & 49.5$\pm$0.2  & 65.4$\pm$0.2  & \textbf{62.3}$\pm$0.2  & \textbf{77.4}$\pm$0.2  \\
\textbf{\textsc{RBF Kernel}} & 42.1$\pm$1.2  & 54.9$\pm$1.1  &46.0$\pm$1.2  & 59.8$\pm$1.0  \\
\textbf{\textsc{RFFs}} (2048d)   & 52.8$\pm$0.9  & 65.4$\pm$0.9  & 61.1$\pm$0.8  & 74.7$\pm$0.9  \\

%\textbf{\textsc{VRFs/\op}} (512 dim) & 53.8$\pm$1.8  & 67.0$\pm$0.9  & 62.3$\pm$0.5  & 74.5 $\pm$0.4  \\

\textbf{\textsc{MetaVRF}} (w/o \lstm, 780d) & 51.3$\pm$0.8  & 66.1$\pm$0.7  & 61.1$\pm$0.7  & 74.3 $\pm$0.9  \\

\textbf{\textsc{MetaVRF}} (vanilla \lstm, 780d) & 53.1$\pm$0.9  & 66.8$\pm$0.7  & 62.1$\pm$0.8  & 76.0$\pm$0.8  \\

\textbf{\textsc{MetaVRF}} (bi-\lstm, 780d) & \textbf{54.2}$\pm$0.8  & \textbf{67.8}$\pm$0.7  & \textbf{63.1}$\pm$0.7  & 76.5$\pm$0.9  \\
\bottomrule
\end{tabular}
% \begin{tablenotes}
% 	\footnotesize
% 	\center{$^*$training with $20$ ways, test on $5$ ways.}
% \end{tablenotes}
%\end{footnotesize}
\end{table*}

\begin{table*}[t]
\small
\caption{Performance ($\%$) on Omniglot.}\label{tab:omniglot}
\centering
\begin{tabular}{lcccc}
\toprule
%\abovespace
& \multicolumn{2}{c}{\textbf{Omniglot, 5-way}} & \multicolumn{2}{c}{\textbf{Omniglot, 20-way}} \\
% \belowspace
\textbf{Method} & 1-shot & 5-shot & 1-shot & 5-shot \\
\midrule
%\abovespace
\textbf{\textsc{Siamese net}}~\citep{koch2015siamese} & 96.7  & 98.4  & 88  & 96.5  \\
\textbf{\textsc{Matching net}}~\citep{vinyals2016matching} & 98.1  & 98.9  & 93.8  & 98.5  \\
\textbf{\textsc{MAML}}~\citep{finn2017model} & 98.7$\pm0.4$  & \textbf{99.9}$\pm$0.1  & 95.8$\pm$0.3  & 98.9$\pm$0.2  \\
\textbf{\textsc{Proto net}}~\citep{snell2017prototypical} & 98.5$\pm$0.2  & 99.5$\pm$0.1  & 95.3$\pm$0.2  & 98.7$\pm$0.1  \\
\textbf{\textsc{SNAIL}}~\citep{mishra2018simple} & 99.1$\pm$0.2  & 99.8 $\pm$0.1  & 97.6 $\pm$0.3  & \textbf{99.4} $\pm$0.2 \\
\textbf{\textsc{GNN}}~\citep{garcia2018few} & 99.2  & 99.7  & 97.4  & 99.0  \\
% \belowspace
%\textbf{\textsc{GNN}}~\citep{garcia2018few} & \textbf{99.2 } & \textbf{99.7 } & \textbf{97.4 } & 99.0  \\

\textbf{\textsc{VERSA}}~\citep{gordon2018meta} & 99.7$\pm$0.2  & 99.8$\pm$0.1  & 97.7$\pm$0.3  & 98.8$\pm$0.2  \\
\textbf{\textsc{R2-D2}}~\citep{bertinetto2018meta} & 98.6  & 99.7 & 94.7  & 98.9  \\
\textbf{\textsc{IMP}}~\citep{allen2019infinite} & 98.4$\pm$0.3  & 99.5$\pm$0.1 & 95.0$\pm$0.1  & 98.6$\pm$0.1  \\
%\textbf{\textsc{iMAML}}~\citep{rajeswaran2019meta} &99.5$\pm$0.3  &\textbf{99.7}$\pm$0.1 & 96.2$\pm$0.4 & \textbf{99.1}$\pm$0.1  \\
\midrule
% \abovespace
\textbf{\textsc{RBF Kernel}} & 95.5$\pm$0.2  & 99.1$\pm$0.2  &92.8$\pm$0.3  & 97.8$\pm$0.2  \\
\textbf{\textsc{RFFs}} (2048d)  & 99.5$\pm$0.2  & 99.5$\pm$0.2  & 97.2$\pm$0.3  & 98.3$\pm$0.2  \\

%\textbf{\textsc{VRFs/\op}} (512 dim) & \textbf{99.6}$\pm$0.3  & \textbf{99.7}$\pm$0.1  & 96.7$\pm$0.3  & 98.7$\pm$0.2  \\
% \belowspace
\textbf{\textsc{MetaVRF}} (w/o \lstm, 780d) & 99.6$\pm$0.2  & 99.6$\pm$0.2  & 97.0$\pm$0.3   & 98.4$\pm$0.2  \\
\textbf{\textsc{MetaVRF}} (vanilla \lstm, 780d) & 99.7$\pm$0.2  & 99.8$\pm$0.1  & 97.5$\pm$0.3   & 99.0$\pm$0.2  \\
\textbf{\textsc{MetaVRF}} (bi-\lstm, 780d) & \textbf{99.8}$\pm$0.1  & \textbf{99.9}$\pm$0.1  & \textbf{97.8}$\pm$0.3   & 99.2$\pm$0.2  \\
\bottomrule
\end{tabular}
\end{table*}

\subsection{Few-Shot Regression}
We conduct regression tasks with different numbers of shots $k$, and compare our MetaVRF with MAML~\citep{finn2017model}, a representative meta-learning algorithm. We follow the MAML work \citep{finn2017model} to fit a target sine function $y=A \sin{(wx + b)}$, with only a few annotated samples. $A \in [0.1, 5]$, $w \in [0.8, 1.2]$, and $ b\in [0, \pi ]$ denote the amplitude, frequency, and phase, respectively, which follow a uniform distribution within the corresponding interval. The goal is to estimate the target sine function given only $n$ randomly sampled data points. In our experiments, we consider the input in the range of $x\in [-5, 5]$, and conduct three tests under the conditions of $k = 3, 5, 10$. For a fair comparison, we compute the feature embedding using a small multi-layer perception (MLP) with two hidden layers of size $40$, following the same settings used in MAML. 

The results in Figure~\ref{fig:reg} show that our MetaVRF fits the function well with only three shots but performs better with an increasing number of shots, almost entirely fitting the target function with ten shots. Moreover, the results demonstrate the advantage of exploring task dependency by LSTM-based inference. MetaVRF with bi-\lstm~performs better than regular LSTM since more context tasks are incorporated by bi-\lstm. In addition, we observe that MetaVRF performs  better than MAML for all three settings with varying numbers of shots. We provide more results on few-shot regression tasks in the supplementary material.

\subsection{Few-Shot Classification}
The classification experiments are conducted on three commonly-used benchmark datasets, i.e., Omniglot \citep{lake2015human}, miniImageNet \citep{vinyals2016matching} and CIFAR-FS \citep{krizhevsky2009learning}; for more details, please refer to the supplementary material. We extract image features using a shallow convolutional neural network with the same architecture as in~\citep{gordon2018meta}. We do not use any fully connected layers for these CNNs. The dimension of all feature vectors is $256$. We also evaluate the random Fourier features (RFFs) and the radial basis function (RBF) kernel in which we take the bandwidth $\sigma$ as the mean of pair-wise distances between samples in the support set of each task. The inference network $\phi(\cdot)$ is a three-layer MLP with $256$ units in the hidden layers and rectifier non-linearity where input sizes are $256$ and $512$ for the vanilla and bidirectional LSTMs, respectively.

The key hyperparameter for the number of bases $D$ in (\ref{rfs}) is set to $D=780$ for MetaVRF in all experiments, while we use RFFs with $D=2048$ as this produces the best performance. The sampling rate in our MetaVRF is much lower than in previous works using RFFs, in which $D$ is usually set to be $5$ to $10$ times the dimension of the input features~\citep{yu2016orthogonal, rahimi2008random}. We adopt a similar meta-testing protocol as~\citep{gordon2018meta, finn2017model}, but we test on $3,000$ episodes rather than $600$ and present the results with $95\%$ confidence intervals. All reported results are produced by models trained from scratch. We compare with previous methods that use the same training procedures and similar shallow conventional CNN architectures as ours. The comparison results on three benchmark datasets are reported in Tables~\ref{tab:miniandcifar} and \ref{tab:omniglot}. %The compared methods use the same shallow CNN architectures. 

%On the Omniglot dataset, previous methods did not specify the splits for training, validation, and testing, which may result in unfair comparisons. Our MetaVRF method falls within the error bars of the state-of-the-art models on all experiments under $5$-way $1$-shot, $5$-way $5$-shot, $20$-way $1$-shot and $20$-way $5$-shot settings. We compare with previous methods that use the same training procedures and conventional CNN architectures as ours. 

%In ~\citet{vinyals2016matching, snell2017prototypical, ravi2017optimization, finn2017model, sung2018learning}, $64$ filters are in each convolutional layer, \citet{ravi2017optimization,finn2017model} set the number of filters to $32$ for \mini ~to avoid overfitting. 

On all benchmark datasets, MetaVRF delivers the state-of-the-art performance. Even with a relatively low sampling rate, MetaVRF produces consistently better performance compared with RBF kernels, RFFs. MetaVRF with bi-\lstm~outperforms the one with vanilla \lstm~since it can leverage more information. It is worth mentioning that MetaVRF with bi-\lstm~achieves good performance ($54.2\%$) under the $5$-way $1$-shot setting on the \mini~dataset, surpassing the second best model by $1\%$. The MetaVRFs with bi-\lstm~and vanilla \lstm~consistently outperform the one without the \lstm, which demonstrates the effectiveness of using \lstm~to explore task dependency. The much better performance over RBF kernels on all three datasets indicates the great benefit of adaptive kernels based on random Fourier features. The relatively inferior performance produced by RBF kernels is due to that the mean of pair-wise distances of support samples would not be able to provide a proper estimate of the kernel bandwidth since we only have a few samples in each task, for instance, five samples under the 5-way 1-shot setting in the support set. 

Note that on Omniglot, the performance of existing methods saturates and MetaVRF with~bi-\lstm~achieves the best performance for most settings, including $5$-way $1$-shot, $5$-way $5$-shot, and $20$-way $1$-shot. It is also competitive under the $20$-way $5$-shot setting falling within the error bars of the state-of-the-arts. In Table~\ref{tab:miniandcifar}, we also implement a MAML ($64C$) with $64$ channels in each convolutional layer. However, while it obtains modest performance, we believe the increased model size leads to overfitting. Since in the original SNAIL, a very deep ResNet-12 network is used for embedding, we cite the result of SNAIL reported in \citet{bertinetto2018meta} using similar shallow networks as ours. For fair comparison, we also cite the original results of R2-D2~\citep{bertinetto2018meta} using $64$ channels.

\begin{table}[t]
\caption{Performance ($\%$) on \mini{} (5-way)}\label{tab:mini}
\small
\centering
\begin{tabular}{lcc}
\toprule
% \abovespace
% %& \multicolumn{2}{c}{\textbf{\mini{}, 5-way}} \\
% \belowspace
\textbf{Method} & 1-shot & 5-shot \\
\midrule
% \abovespace
%\textbf{\textsc{Matching net}}~\citep{vinyals2016matching} & 41.2  & 56.2  & --- & --- \\
\textbf{\textsc{Meta-SGD}}~\citep{li2017meta} & 54.24$\pm$0.03& 70.86$\pm$0.04  \\
\textbf{\textsc{}}\citep{gidaris2018dynamic}& 56.20$\pm$0.86& 73.00$\pm$0.64  \\
\textbf{\textsc{}}\citep{bauer2017discriminative}& 56.30$\pm$0.40& 73.90$\pm$0.30  \\
\textbf{\textsc{}}\citep{munkhdalai2017rapid}& 57.10$\pm$0.70& 70.04$\pm$0.63  \\
\textbf{\textsc{}}\citep{qiao2018few}& 59.60$\pm$0.41& 73.54$\pm$0.19  \\
\textbf{\textsc{LEO}}~\citep{rusu2018meta} & 61.76$\pm$0.08& 77.59$\pm$0.12  \\
\textbf{\textsc{SNAIL}}~\citep{mishra2018simple} & 55.71$\pm$0.99& 68.88$\pm$0.92  \\
\textbf{\textsc{TADAM}}~\citep{oreshkin2018tadam} & 58.50$\pm$0.30& 76.70$\pm$0.30  \\
\hline
\textbf{\textsc{MetaVRF}} (w/o \lstm, 780d)  & \textbf{62.12}$\pm$0.07  & 77.05$\pm$0.28 \\
\textbf{\textsc{MetaVRF}}  (vanilla \lstm, 780d) & \textbf{63.21}$\pm$0.06  & \textbf{77.83}$\pm$0.28 \\
\textbf{\textsc{MetaVRF}}  (bi-\lstm, 780d) & \textbf{63.80}$\pm$0.05  & \textbf{77.97}$\pm$0.28 \\
\bottomrule
\end{tabular}
\end{table}

\subsection{Further Analysis}
\textbf{Deep embedding.} Our MetaVRF is independent of the convolutional architectures for feature extraction and can work with deeper embeddings either pre-trained or trained from scratch. In general, the performance improves with more powerful feature extraction architectures. We evaluate our method using pre-trained embeddings in order to compare with existing methods using deep embedding architectures.
To benchmark with those methods, we adopt the pre-trained embeddings from a 28-layer wide residual network (WRN-28-10) \citep{zagoruyko2016wide}, in a similar fashion to \citep{rusu2018meta, bauer2017discriminative, qiao2018few}. We choose activations in the 21-st layer, with average pooling over spatial dimensions, as feature embeddings. The dimension of pre-trained embeddings is $640$. We show the comparison results on the \mini~dataset for 5-way 1-shot and 5-shot settings in Table.~\ref{tab:mini}. Our MetaVRF with bi-\lstm~achieves the best performance under both settings and largely surpasses LEO, a recently proposed meta-learning method, especially on the challenging 5-way 1-shot setting. Note that the MetaVRF with vanilla \lstm~and without \lstm~also produce competitive performance.

\begin{figure}[t]
	\centering
	\begin{subfigure}%{0.4\columnwidth}
		\centering
		\includegraphics[width=0.49\columnwidth]{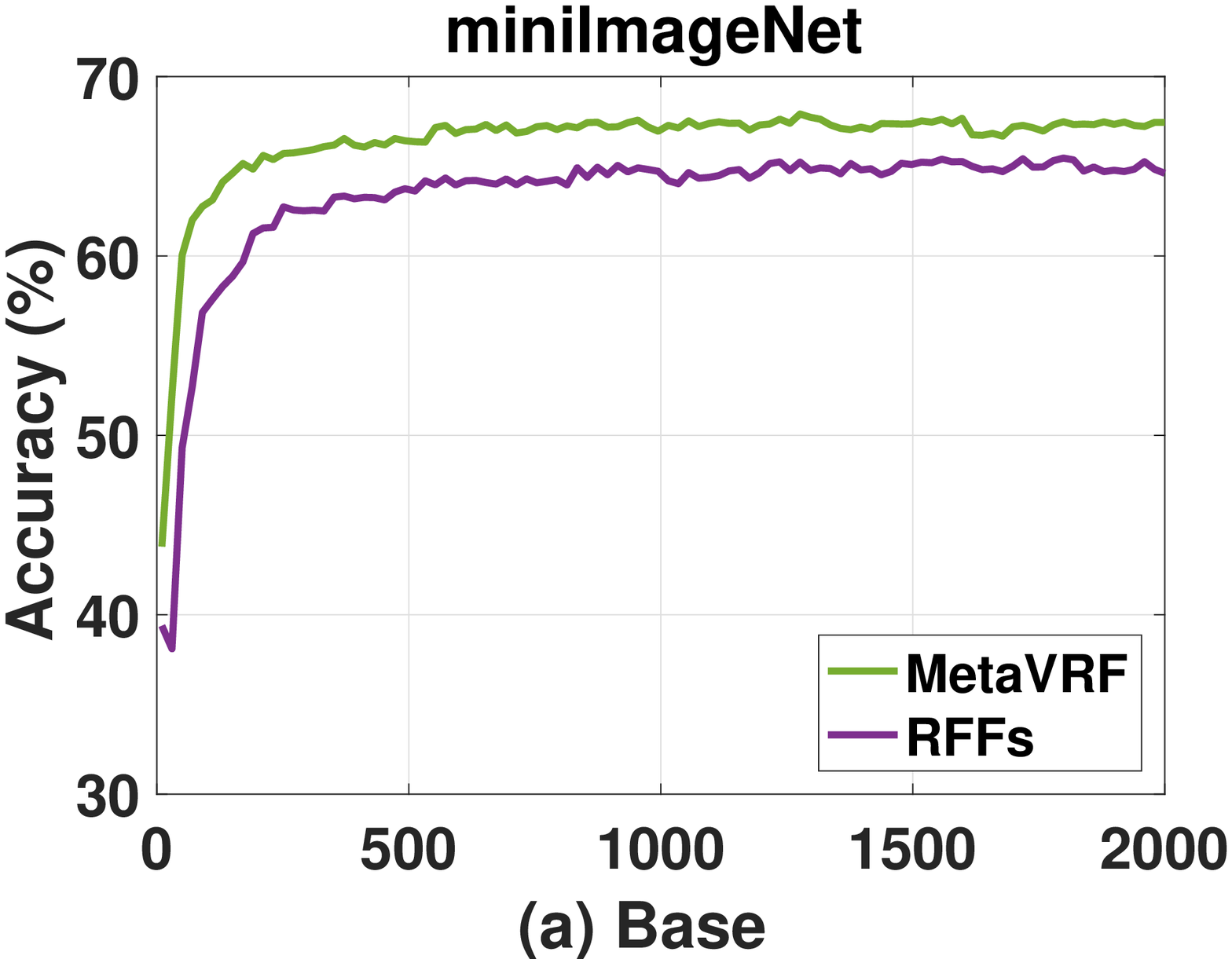}
		\label{fig:eff1}
		%\caption{\label{fig:flex1} .}
	\end{subfigure}%
	%~ 
	\begin{subfigure}%{0.4\columnwidth}
		\centering
		\includegraphics[width=0.49\columnwidth]{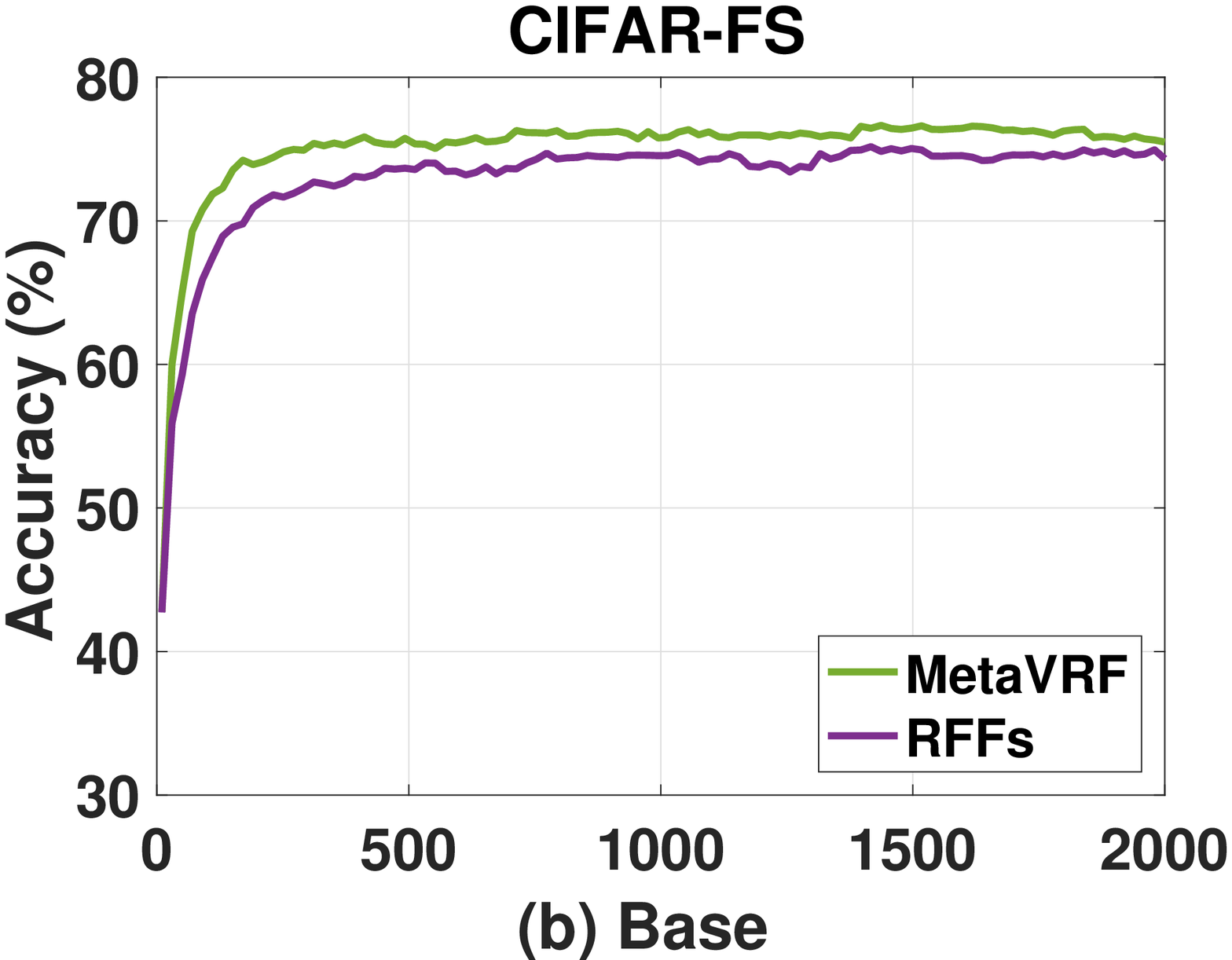}
		\label{fig:eff2}
		%\caption{\label{fig:flex2} .}
	\end{subfigure}
    \vspace{-5mm}
	\caption{ Performance with different numbers $D$ of bases. Our MetaVRF consistently achieves better performance than regular RFFs, especially with relatively low sampling rates.}
	\label{fig:eff} 
\end{figure}

\textbf{Efficiency.} Regular random Fourier features (RFFs) usually require high sampling rates to achieve satisfactory performance. However, our MetaVRF can achieve high performance with a relatively low sampling rate compared, which guarantees its high efficiency. In Figure~\ref{fig:eff}, we compare with regular RFFs using different sampling rates. We show the performance change of fully trained models using RFFs and our MetaVRF with bi-\lstm~under a different number $D$ of bases. We show the comparison results for the $5$-way $5$-shot setting in Figure~\ref{fig:eff}. MetaVRF with bi-\lstm~consistently yields higher performance than regular RFFs with the same number of sampled bases. The results verify the efficiency of our MetaVRF in learning adaptive kernels and the effectiveness in improving performance by exploring dependencies of related tasks.

\begin{figure}[]
	\centering
	\begin{subfigure}%{0.4\columnwidth}
		\centering
		\includegraphics[width=0.49\columnwidth]{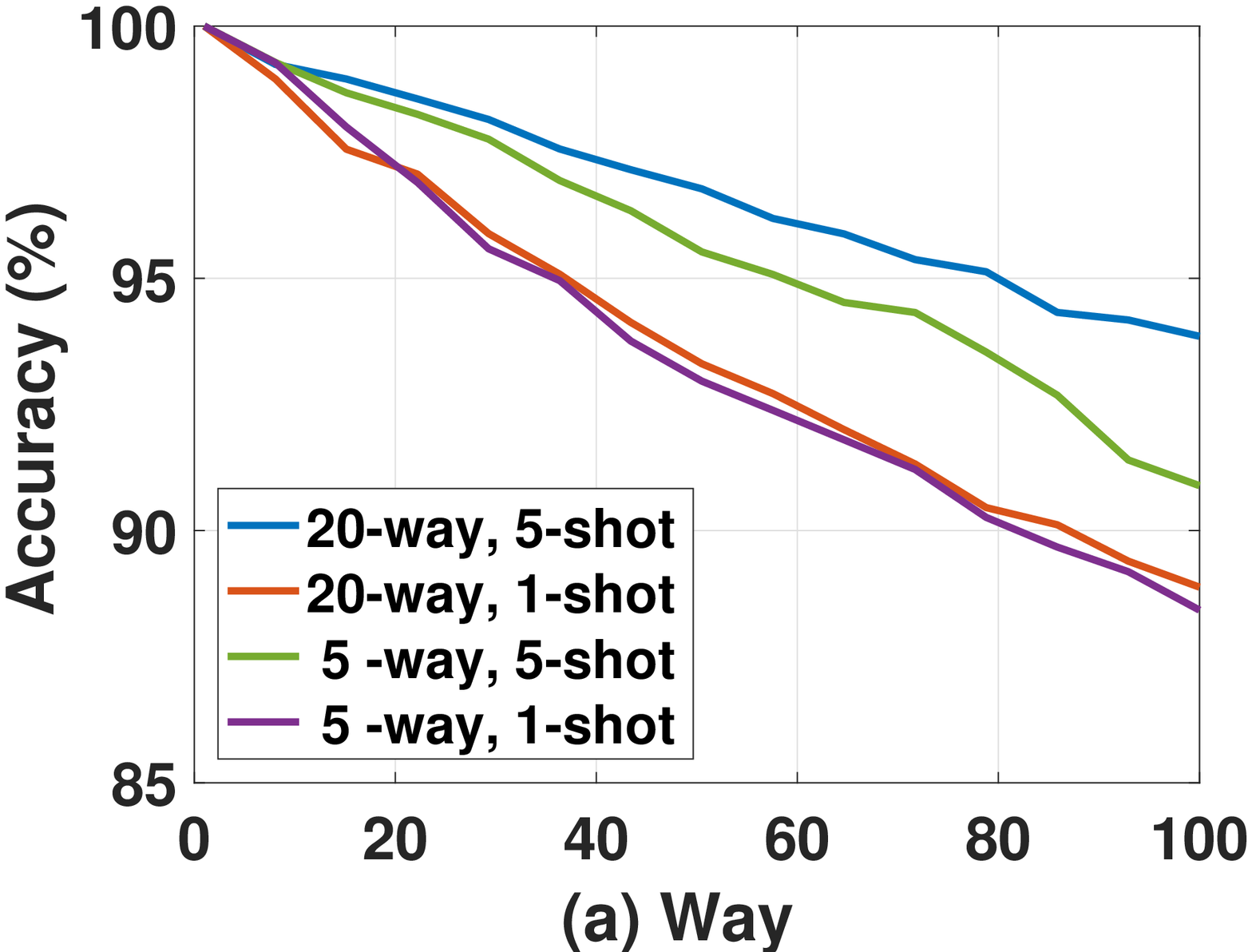}
		%\caption{\label{fig:flex1} (a) number of way}
	\end{subfigure}%
	\begin{subfigure}%{0.4\columnwidth}
		\centering
		\includegraphics[width=0.49\columnwidth]{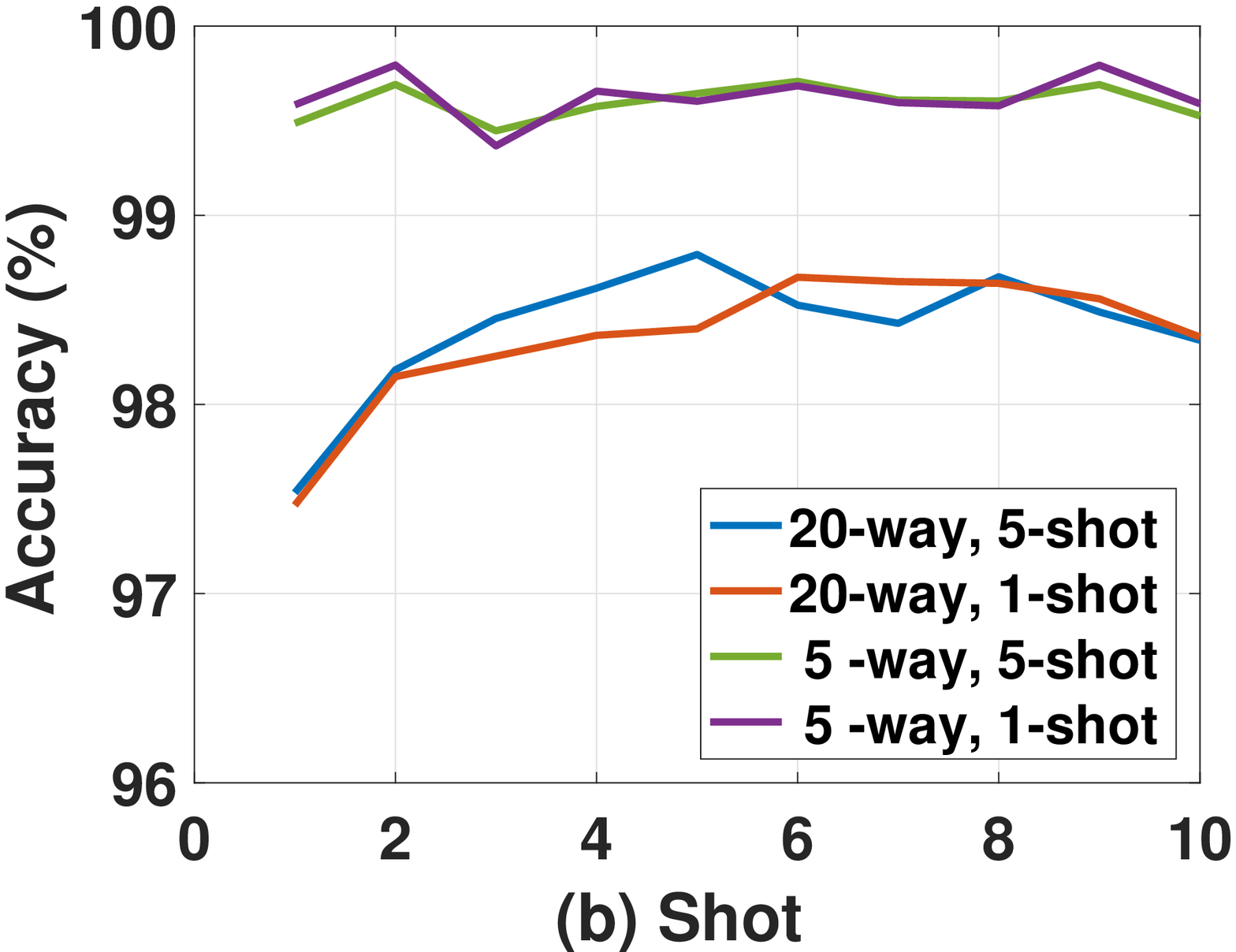}
		%\caption{(b) umber of shot.}
	\end{subfigure}
	    \vspace{-5mm}
	%\caption{Performance on test tasks with varied ways and shots on Omniglot.}
	\caption{ Performance with varied ways and shots on Omniglot.}
	\label{fig:flex}
\end{figure}

\textbf{Versatility.} In contrast to most existing meta-learning methods, our MetaVRF can be used for versatile settings.
We evaluate the performance of MetaVRF on more challenging scenarios where the number of ways $C$ and  shots $k$ between training and testing are inconsistent. Specifically, we test the performance of MetaVRF on tasks with varied $C$ and $k$, when it is trained on one particular $C$-way-$k$-shot task. As shown in Figure~\ref{fig:flex}, the results demonstrate that the trained model can still produce good performance, even on the challenging condition with a far higher number of ways. In particular, the model trained on the $20$-way-$5$-shot task can retain a high accuracy of $94\%$ on the $100$-way setting, as shown in Figure~\ref{fig:flex}(a). The results also indicate that our MetaVRF exhibits considerable robustness and flexibility to a great variety of testing conditions.

\section{Conclusion}
\label{sec:conclusions}
In this paper, we introduce kernel approximation based on random Fourier features into the meta-learning framework for few-shot learning. We propose meta variational random features (MetaVRF), which leverage variational inference and meta-learning to infer the spectral distribution of random Fourier features in a data-driven way. MetaVRF generates random Fourier features of high representational power with a relatively low spectral sampling rate by using an LSTM based inference network to explore the shared knowledge. In practice, our LSTM-based inference network demonstrates a great ability to collect and absorb common knowledge from experience tasks, which enables quick adaptation to specific tasks for improved performance. Experimental results on both regression and classification tasks demonstrate the effectiveness for few-shot learning.

\section*{Acknowledgements}
This research was supported in part by Natural Science Foundation of China (No. 61976060, 61871016, 61876098).

\bibliography{final}
\bibliographystyle{icml2020}

\clearpage

\onecolumn

\section{Appendix}

\subsection{Derivations of the ELBO}
\label{ELBO}
For a singe task, we begin with maximizing log-likelihood of the conditional distribution $p(\mathbf{y} | \mathbf{x}, \mathcal{S})$ to derive the ELBO of MetaVRF. By leveraging Jensen's inequality, we have the following steps as
\begin{align}
\log  p(\mathbf{y} | \mathbf{x},\mathcal{S})&= \log  \int p(\mathbf{y} | \mathbf{x}, \mathcal{S}, \bm{\omega})  p(\bm{\omega} | \mathbf{x}, \mathcal{S}) d\bm{\omega} \\
&= \log  \int p(\mathbf{y} | \mathbf{x}, \mathcal{S}, \bm{\omega})  p(\bm{\omega} | \mathbf{x}, \mathcal{S}) \frac{q_{\phi}(\bm{\omega}| \mathcal{S})}{q_{\phi}(\bm{\omega}| \mathcal{S})} d\bm{\omega}\\
&\geq \int \log \left[ \frac{ p(\mathbf{y} | \mathbf{x}, \mathcal{S}, \bm{\omega})  p(\bm{\omega} | \mathbf{x}, \mathcal{S}) }{q_{\phi}(\bm{\omega}| \mathcal{S})} \right] q_{\phi}(\bm{\omega}|  \mathcal{S}) d\bm{\omega} \\
&= \underbrace{\mathbb{E}_{q_{\phi}(\bm{\omega}| \mathcal{S})} \log \, [p(\mathbf{y} | \mathbf{x},  \mathcal{S},\bm{\omega} )] - \KL[q_{\phi}(\bm{\omega}|\mathcal{S}) || p(\bm{\omega} | \mathbf{x}, \mathcal{S})]}_{\text{ELBO}}.
\label{der-likeli}
\end{align}

The ELBO  can also be derived from the perspective of the KL divergence between the variational posterior $q_{\phi}(\bm{\omega}| \mathcal{S})$ and the posterior $p(\bm{\omega} | \mathbf{y}, \mathbf{x}, \mathcal{S})$:
\begin{equation}
\begin{aligned}
\KL[q_{\phi}(\bm{\omega}| \mathcal{S}) || p(\bm{\omega} | \mathbf{y}, \mathbf{x}, \mathcal{S})]
& = \mathbb{E}_{q_{\phi}(\bm{\omega}| \mathcal{S})} \left[\log q_{\phi}(\bm{\omega}| \mathcal{S}) - \log p(\bm{\omega} | \mathbf{y}, \mathbf{x}, \mathcal{S})\right]\\
& = \mathbb{E}_{q_{\phi}(\bm{\omega}| \mathcal{S})} \left[\log q_{\phi}(\bm{\omega}| \mathcal{S}) - \log \frac{p(\mathbf{y} | \bm{\omega}, \mathbf{x}, \mathcal{S}) p(\bm{\omega}| \mathbf{x}, \mathcal{S})}{p(\mathbf{y}|  \mathbf{x}, \mathcal{S}) }\right] \\&= \log p(\mathbf{y}| \mathbf{x}, \mathcal{S}) + \mathbb{E}_{q_{\phi}(\bm{\omega}| \mathcal{S})} \left[\log q_{\phi}(\bm{\omega}| \mathcal{S})- \log p(\mathbf{y} | \bm{\omega}, \mathbf{x}, \mathcal{S}) - \log p(\bm{\omega}| \mathbf{x}, \mathcal{S})\right]\\
&= \log p(\mathbf{y}| \mathbf{x}, \mathcal{S}) - \mathbb{E}_{q_{\phi}(\bm{\omega}| \mathcal{S})} \left[ \log p(\mathbf{y} | \bm{\omega}, \mathbf{x}, \mathcal{S})\right] + \KL[ q_{\phi}(\bm{\omega}| \mathcal{S}) || p(\bm{\omega}| \mathbf{x}, \mathcal{S})] \geq 0.
\label{der-kl}
\end{aligned}
\end{equation}

Therefore, the lower bound of the $\log p(\mathbf{y}| \mathbf{x}, 
\mathcal{S})$ is 
\begin{equation}
\begin{aligned}
\log  p(\mathbf{y} | \mathbf{x}, \mathcal{S}) &\geq \mathbb{E}_{q_{\phi}(\bm{\omega}| \mathcal{S})} \log \, [p(\mathbf{y} | \mathbf{x}, \mathcal{S}, \bm{\omega} )]  - \KL[q_{\phi}(\bm{\omega}|\mathcal{S}) || p(\bm{\omega} | \mathbf{x}, \mathcal{S})],
\label{elbo}
\end{aligned}
\end{equation}
which is consistent with (\ref{der-likeli}).

\subsection{Cross attention in the prior network}

In $p(\bm{\omega} | \mathbf{x}, \mathcal{S})$, both $\mathbf{x}$ and $\mathcal{S}$ are inputs of the prior network. In order to effectively integrate the two conditions, we adopt the cross attention \citep{kim2019attentive} between $\mathbf{x}$ and each element in $\mathcal{S}$. In our case, we have the key-value matrices $K = V \in \mathbb{R}^{C \times d} $, where $d$ is the dimension of the feature representation, and $C$ is the number of categories in the support set. We adopt the instance pooling by taking the average of samples in each category when the shot number $k>1$.

For the query $Q_i = \mathbf{x} \in \mathbb{R}^{d} $, the Laplace kernel returns attentive representation for $\mathbf{x}$:
\begin{equation}
\begin{aligned}
\textbf{Laplace}&(Q_i,K,V) := W_iV \in \mathbb{R}^{d}, \quad W_i := \softmax(-\left\|Q_i-K_{j.} \right\|_{1})^C_{j=1}
\label{Laplace-att}
\end{aligned}
\end{equation}
The prior network takes the attentive representation as the input.

\iffalse
\subsection{Experimental results with regular RBF kernels}
\label{rbf}

To evaluate the effectiveness of learning the kernel with random features, we conduct the experiment of a fixed kernel (\textit{e.g.} the RBF kernel). We adopt the same training-testing strategy in the meta-learning scenario. The bandwidth is set as the mean of all pairwise distances. As shown in Table \ref{tab:rbf}, MetaVRF is greatly superior to the fixed RBF kernel for few-shot classification. This result indicates the kernel learned by our model with random features shows favorable scalability and avoids vulnerable heuristic bandwidth tuning. 

\begin{table*}
	\small
	\caption{Performance ($\%$) on \mini{} and \cifarfs{}.}\label{tab:rbf}
	\centering
	%\begin{center}
	%\begin{footnotesize}
	%\begin{scriptsize}
	\begin{tabular}{lcccc}
		%\vspace{-0.05in}
		\toprule
		% \abovespace
		& \multicolumn{2}{c}{\textbf{\mini{}, 5-way}} & \multicolumn{2}{c}{\textbf{\cifarfs{}, 5-way}} \\
		% \belowspace
		\textbf{Method} & 1-shot & 5-shot & 1-shot & 5-shot \\
		\midrule
		
		\textbf{\textsc{RBF}} & 42.1$\pm$1.2  & 54.9$\pm$1.1  &46.0$\pm$1.2  & 59.8$\pm$1.0  \\
		
		\textbf{\textsc{MetaVRF}} (bi-\lstm, 780d) & \textbf{54.2}$\pm$0.8  & \textbf{67.8}$\pm$0.7  & \textbf{63.1}$\pm$0.7  & \textbf{76.5}$\pm$0.9  \\
		\bottomrule
	\end{tabular}
\end{table*}

\fi

\subsection{More experimental details}
\label{med}
We train all models using the Adam optimizer~\citep{kingma2014adam} with a learning rate of $0.0001$.\ The other training setting and network architecture for regression and classification on three datasets are different as follows.

\subsubsection{Inference networks}
The architecture of the inference network with vanilla LSTM for the regression task is in Table \ref{inference-regression-1}. The architecture of the inference network with  bidirectional LSTM for the regression task is in Table \ref{inference-regression-2}. For few-shot classification tasks, all models share the same architecture with vanilla LSTM, as in Table \ref{inference-network-1}, For few-shot classification tasks, all models share the same architecture with  bidirectional LSTM, as in Table \ref{inference-network-2}.

\subsubsection{Prior networks}
The architecture of the prior network for the regression task is in Table \ref{prior-regression}. For few-shot classification tasks, all models share the same architecture, as in Table \ref{prior-network}.

\subsubsection{Feature embedding networks}

\textbf{Regression.}
The fully connected architecture for regression tasks is shown in Table \ref{fcn-reg}. We train all three models ($3$-shot, $5$-shot, $10$-shot) over a total of $20,000$ iterations, with $6$ episodes per iteration.

\textbf{Classification.}
The CNN architectures for Omniglot, \cifarfs{}, and \mini{} are shown in Table \ref{cnn-Omn}, \ref{cnn-cifar}, and \ref{cnn-mini}. The difference of feature embedding architectures for different datasets is due the different image sizes.

\begin{table}
	\small
	\begin{center}
		\caption{The inference network $\phi(\cdot)$ based on the vanilla LSTM used for regression.}
		\centering
		\begin{tabular}{cl}
			\toprule
			\textbf{Output size} & \textbf{Layers} \\
			\midrule
			$40$ & Input samples feature \\
			$40$ & fully connected, ELU \\
			$40$ & fully connected, ELU \\
			$40$ &  LSTM cell, Tanh to $\mu_w$, $\log\sigma^2_w$\\
			\bottomrule
		\end{tabular}
		\label{inference-regression-1}
	\end{center}
\end{table}

\begin{table} 
	\small
	\begin{center}
		\caption{The inference network $\phi(\cdot)$ based on the bidirectional LSTM for regression. }
		\centering
		\begin{tabular}{cl}
			\toprule
			\textbf{Output size} & \textbf{Layers} \\
			\midrule
			$80$ & Input samples feature \\
			$40$ & fully connected, ELU \\
			$40$ & fully connected, ELU \\
			$40$ &  LSTM cell, Tanh to $\mu_w$, $\log\sigma^2_w$\\
			
			\bottomrule
		\end{tabular}
		\label{inference-regression-2}
	\end{center}
\end{table}

\begin{table} 
	\small
	\begin{center}
		\caption{The inference network $\phi(\cdot)$ based on the vanilla LSTM for  Omniglot, \mini{}, \cifarfs{}.}
		\centering
		\begin{tabular}{cl}
			\toprule
			\textbf{Output size} & \textbf{Layers} \\
			\midrule
			$k \times 256$ & Input feature \\
			$256$ & instance pooling \\
			$256$ & fully connected, ELU \\
			$256$ & fully connected, ELU \\
			$256$ &  fully connected, ELU\\
			$256$ &  LSTM cell, tanh to $\mu_w$, $\log\sigma^2_w$ \\
			
			\bottomrule
		\end{tabular}
		\label{inference-network-1}
	\end{center}
\end{table}

\begin{table} 
	\small
	\begin{center}
		\caption{The inference network $\phi(\cdot)$ based on the bidirectional LSTM for  Omniglot, \mini{}, \cifarfs{}.}
		\centering
		\begin{tabular}{cl}
			\toprule
			\textbf{Output size} & \textbf{Layers} \\
			\midrule
			$k \times 512$ & Input feature \\
			$256$ & instance pooling \\
			$256$ & fully connected, ELU \\
			$256$ & fully connected, ELU \\
			$256$ &  fully connected, ELU\\
			$256$ &  LSTM cell, tanh to $\mu_w$, $\log\sigma^2_w$ \\
			
			\bottomrule
		\end{tabular}
		\label{inference-network-2}
	\end{center}
\end{table}

\begin{table} 
	\small
	\begin{center}
		\caption{The prior network for regression.}
		\centering
		\begin{tabular}{cl}
			\toprule
			\textbf{Output size} & \textbf{Layers} \\
			\midrule
			$40$ & fully connected, ELU \\
			$40$ & fully connected, ELU \\
			$40$ &  fully connected  to $\mu_w$, $\log\sigma^2_w$ \\
			
			\bottomrule
		\end{tabular}
		\label{prior-regression}
	\end{center}
\end{table}

\begin{table} 
	\small
	\begin{center}
		\caption{The prior network for Omniglot, \mini{}, \cifarfs{}}
		\centering
		\begin{tabular}{cl}
			\toprule
			\textbf{Output size} & \textbf{Layers} \\
			\midrule
			$256$ & Input query feature\\
			$256$ & fully connected, ELU \\
			$256$ & fully connected, ELU \\
			$256$ &  fully connected  to $\mu_w$, $\log\sigma^2_w$ \\
			
			\bottomrule
		\end{tabular}
		\label{prior-network}
	\end{center}
\end{table}

\subsection{Few-shot classification datasets}
\label{fscd}

\textbf{Omniglot}~\citep{lake2015human} is a benchmark of few-shot learning that contain $1623$ handwritten characters (each with $20$ examples). All characters are grouped in $50$ alphabets.
For fair comparison against the state of the arts, we follow the same data split and pre-processing used in ~\citet{vinyals2016matching}. The training, validation, and testing are composed of a random split of $[1100, 200, 423]$. The dataset is augmented with rotations of $90$ degrees, which results in $4000$ classes for training, $400$ for validation, and $1292$ for testing. The number of examples is fixed as $20$. All images are resized to $28\mathord\times 28$. For a $C$-way, $k$-shot task at training time, we randomly sample $C$ classes from the $4000$ classes. Once we have $C$ classes, $(k+15)$ examples of each are sampled. Thus, there are $C\mathord\times k$ examples in the support set and $C \mathord\times 15$ examples in the query set. The same sampling strategy is also used in validation and testing.

\textbf{\mini{}}~\citep{vinyals2016matching} is a challenging dataset constructed from ImageNet~\citep{russakovsky2015imagenet}, which comprises a total of  $100$ different classes (each with $600$ instances). All these images have been downsampled to $84\mathord\times 84$.
We use the same splits of~\citet{ravi2017optimization}, where there are $[64, 16, 20]$ classes for training, validation and testing.  We use the same episodic manner as Omniglot for sampling.

\textbf{\cifarfs{}} (CIFAR100 few-shots)~\citep{bertinetto2018meta} is adapted from the CIFAR-100 dataset~\citep{krizhevsky2009learning} for few-shot learning.  Recall that in the image classification benchmark CIFAR-100, there are $100$ classes grouped into $20$ superclasses (each with $600$ instances). \cifarfs{} use the same split criteria ($64, 16, 20$) with which \mini{} has been generated. The resolution of all images is $32\mathord\times 32$.

\begin{table} 
	\small
	\begin{center}
		\caption{The fully connected network $\psi(\cdot)$ used for regression.}
		\centering
		\begin{tabular}{cl}
			\toprule
			\textbf{Output size} & \textbf{Layers} \\
			\midrule
			$1$ & Input training samples \\
			$40$ & fully connected, RELU \\
			$40$ & fully connected, RELU \\
			\bottomrule
		\end{tabular}
		\label{fcn-reg}
	\end{center}
\end{table}

\begin{table*} 
	\small
	\begin{center}
		\caption{The CNN architecture $\psi(\cdot)$ for Omniglot.}
		\begin{tabular}{p{1.7cm}  p{11.5cm} }
			\hline
			Output size    & Layers \\
			\hline
			$ 28 \mathord\times 28 \mathord\times 1$  &Input images\\\hline
			$ 14 \mathord\times 14 \mathord\times 64$  & \textit{conv2d} ($3 \mathord\times 3$, stride=1, SAME, RELU), dropout 0.9, \textit{pool} ($2 \mathord\times 2$, stride=2, SAME)\\ 
			$ 7 \mathord\times 7 \mathord\times 64$  & \textit{conv2d} ($3 \mathord\times 3$, stride=1, SAME, RELU), dropout 0.9, \textit{pool} ($2 \mathord\times 2$, stride=2, SAME)\\ 
			$ 4 \mathord\times 4 \mathord\times 64$  & \textit{conv2d} ($3 \mathord\times 3$, stride=1, SAME, RELU), dropout 0.9, \textit{pool} ($2 \mathord\times 2$, stride=2, SAME)\\ 
			$ 2 \mathord\times 2 \mathord\times 64$  & \textit{conv2d} ($3 \mathord\times 3$, stride=1, SAME, RELU), dropout 0.9, \textit{pool} ($2 \mathord\times 2$, stride=2, SAME)\\ 
			256 & flatten\\
			\hline
		\end{tabular}
		\label{cnn-Omn}
	\end{center}
\end{table*}

\begin{table*} 
	\small
	\begin{center}
		\caption{The CNN architecture $\psi(\cdot)$  for \cifarfs{}}
		\begin{tabular}{p{1.7cm}  p{11.5cm} }
			\hline
			Output size    & Layers \\
			\hline
			$ 32 \mathord\times 32 \mathord\times 3$  &Input images\\\hline
			$ 16 \mathord\times 16 \mathord\times 64$  & \textit{conv2d} ($3 \mathord\times 3$, stride=1, SAME, RELU), dropout 0.5, \textit{pool} ($2 \mathord\times 2$, stride=2, SAME)\\ 
			$ 8 \mathord\times 8 \mathord\times 64$  & \textit{conv2d} ($3 \mathord\times 3$, stride=1, SAME, RELU), dropout 0.5, \textit{pool} ($2 \mathord\times 2$, stride=2, SAME)\\ 
			$ 4 \mathord\times 4 \mathord\times 64$  & \textit{conv2d} ($3 \mathord\times 3$, stride=1, SAME, RELU), dropout 0.5, \textit{pool} ($2 \mathord\times 2$, stride=2, SAME)\\ 
			$ 2 \mathord\times 2 \mathord\times 64$  & \textit{conv2d} ($3 \mathord\times 3$, stride=1, SAME, RELU), dropout 0.5, \textit{pool} ($2 \mathord\times 2$, stride=2, SAME)\\ 
			256 & flatten\\
			\hline
		\end{tabular}
		\label{cnn-cifar}
	\end{center}
\end{table*}

\begin{table*} 
	\small
	\begin{center}
		\caption{The CNN architecture $\psi(\cdot)$ for \mini{}}
		\begin{tabular}{p{1.7cm}  p{11.5cm} }
			\hline
			Output size    & Layers \\
			\hline
			$ 84 \mathord\times 84 \mathord\times 3$  &Input images\\\hline
			$ 42 \mathord\times 42 \mathord\times 64$  & \textit{conv2d} ($3 \mathord\times 3$, stride=1, SAME, RELU), dropout 0.5, \textit{pool} ($2 \mathord\times 2$, stride=2, SAME)\\ 
			$ 21 \mathord\times 21 \mathord\times 64$  & \textit{conv2d} ($3 \mathord\times 3$, stride=1, SAME, RELU), dropout 0.5, \textit{pool} ($2 \mathord\times 2$, stride=2, SAME)\\ 
			$ 10 \mathord\times 10 \mathord\times 64$  & \textit{conv2d} ($3 \mathord\times 3$, stride=1, SAME, RELU), dropout 0.5, \textit{pool} ($2 \mathord\times 2$, stride=2, SAME)\\ 
			$ 5 \mathord\times 5 \mathord\times 64$  & \textit{conv2d} ($3 \mathord\times 3$, stride=1, SAME, RELU), dropout 0.5, \textit{pool} ($2 \mathord\times 2$, stride=2, SAME)\\ 
			$ 2 \mathord\times 2 \mathord\times 64$  & \textit{conv2d} ($3 \mathord\times 3$, stride=1, SAME, RELU), dropout 0.5, \textit{pool} ($2 \mathord\times 2$, stride=2, SAME)\\ 
			256 & flatten\\
			\hline
		\end{tabular}
		\label{cnn-mini}
	\end{center}
\end{table*}

\subsubsection{Other settings}
The settings including the iteration numbers and the batch sizes are different on different datasets. The detailed information is given in Table \ref{opt}.

\begin{table} 
	\small
	\begin{center}
		\caption{The iteration numbers and batch sizes on different datasets.}
		\centering
		\begin{tabular}{lrr}
			\toprule
			\textbf{Dataset} & \textbf{Iteration} & \textbf{Batch size} \\
			\midrule
			Regression & $20,000$ & $25$ \\
			Omniglot & $100,000$ & $6$ \\
			\cifarfs{} & $200,000$ & $8$ \\
			\mini &  $150,000$ & $8$ \\
			\bottomrule
		\end{tabular}
		\label{opt}
	\end{center}
\end{table}

\subsection{More results on few-shot regression}
\label{fsr}
\begin{figure*} 
	\centering
	\begin{subfigure}
		\centering
		\includegraphics[width=0.75\linewidth]{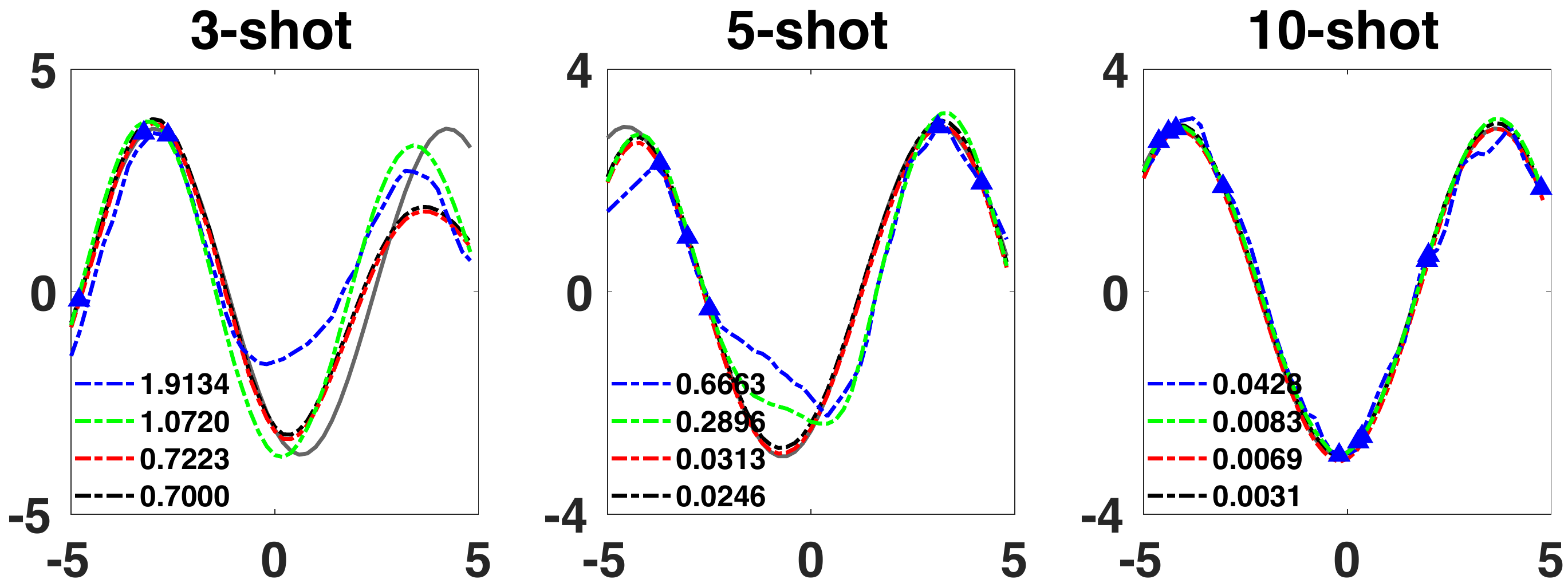}
	\end{subfigure}
	
	\begin{subfigure}
		\centering
		\includegraphics[width=0.75\linewidth]{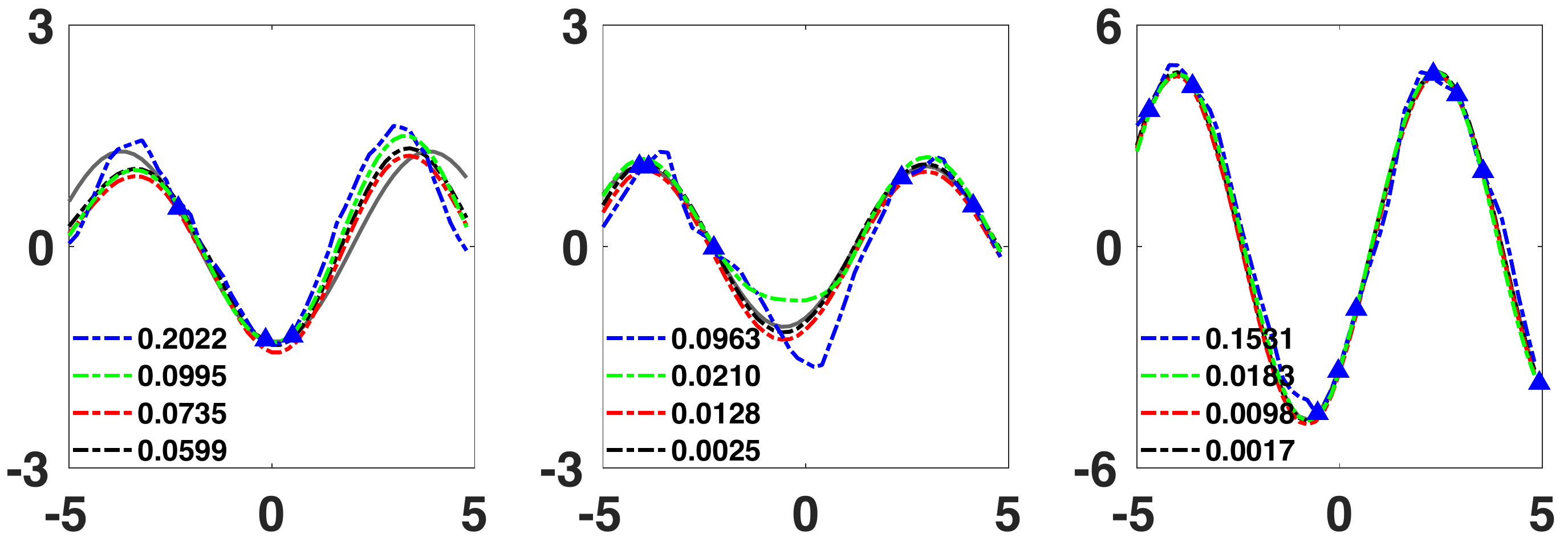}
	\end{subfigure}
	
	\begin{subfigure}
		\centering
		\includegraphics[width=0.75\linewidth]{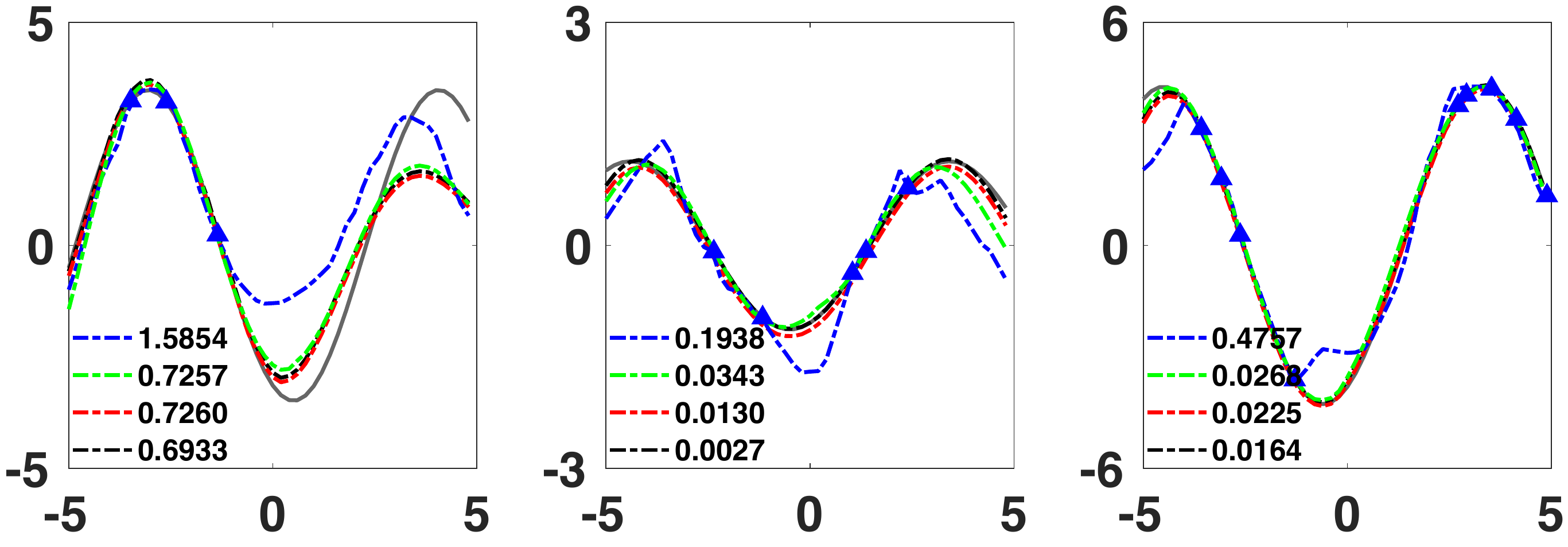}
	\end{subfigure}
	
	\begin{subfigure}
		\centering
		\includegraphics[width=0.75\linewidth]{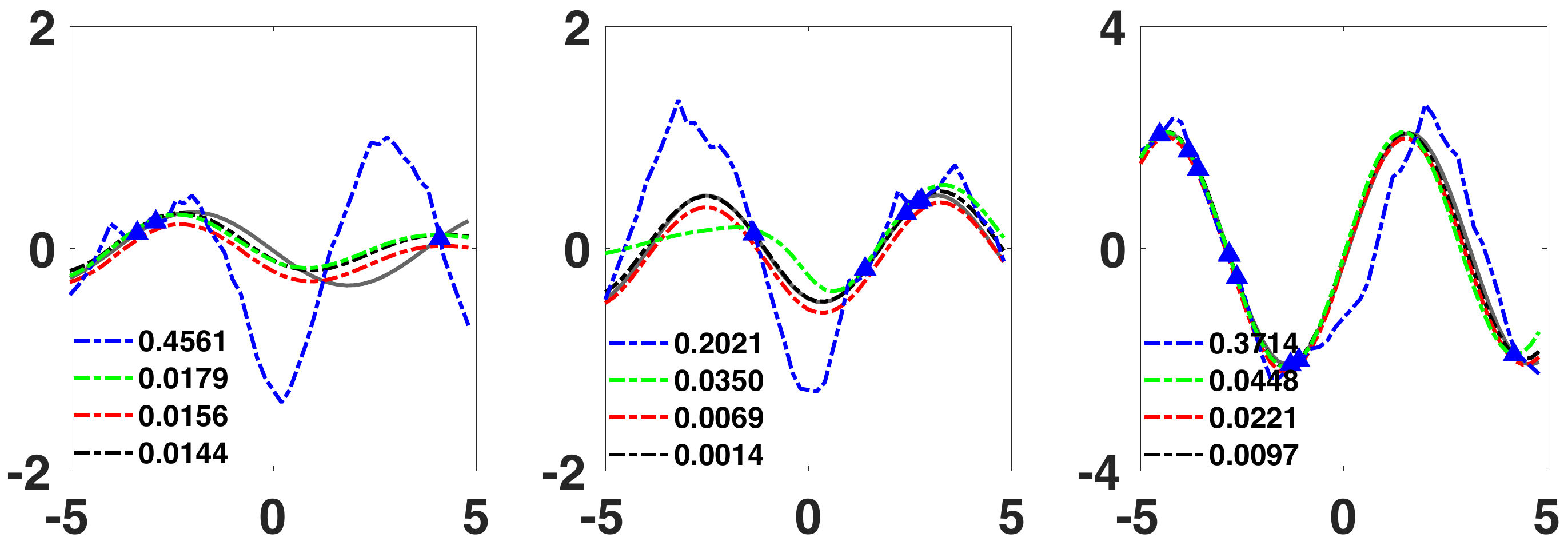}
	\end{subfigure}
	
	% 		\begin{subfigure}
	% 		\centering
	% 		\includegraphics[width=0.9\linewidth]{./figs/r_1.pdf}
	% 	\end{subfigure}
	\caption{\label{afig:reg-ap} More results of few-shot regression.
		\scriptsize{(\blackline~MetaVRF~with~bi-\lstm; \redline~MetaVRF~with~\lstm; \greenline  MetaVRF w/o \lstm; \blueline~MAML; \grayline~Ground Truth; \purplerectangle~Support Samples.})}
	\label{morefsr}
\end{figure*}

We provide more experimental results for the tasks of few-shot regression in Figure~\ref{morefsr}. The proposed MetaVRF again performs much better than regular random Fourier features (RFFs) and the MAML method.

\end{document}